\documentclass{ecai}

\usepackage{latexsym}
\usepackage{amssymb}
\usepackage{amsmath}
\usepackage{amsthm}
\usepackage{booktabs}
\usepackage{enumitem}
\usepackage{graphicx}
\usepackage{color}

\newcommand{\BibTeX}{B\kern-.05em{\sc i\kern-.025em b}\kern-.08em\TeX}

\usepackage[svgnames, table]{xcolor}

\newcommand{\vect}[1]{\ensuremath{\bm{#1}}}

\usepackage[tight-spacing=true]{siunitx} %
\usepackage[export]{adjustbox}
\usepackage{lscape}
\usepackage{makecell}
\usepackage{xspace}
\usepackage{fwt_math_and_notation}
\usepackage{fwt_lp}
\usepackage{cuted}
\usepackage{framed}
\usepackage{bm}

\usepackage{tikz-network}
\usepackage{tikz}
\usetikzlibrary{fit, calc}
\usetikzlibrary{shapes.multipart}  %

\usepackage[ruled,vlined,linesnumbered, commentsnumbered]{algorithm2e} %
\SetKwProg{function}{Function}{}{}
\SetKwProg{class}{Class}{}{}
\SetKwRepeat{ForIUntil}{for \(i \in \{0,1,\dots\}\)}{until}

\newcommand{\bestPerformer}{\bfseries}
\robustify{\bfseries}
\newcommand{\bestPerformerCROC}{}%
\newcommand{\bestPerformerLMCut}{}%

\usepackage{enumerate}

\usepackage{multirow}

\usepackage{cleveref}
\crefname{algorithm}{alg.}{algs.}
\crefname{line}{line}{lines}
\crefname{def}{defn.}{defns.}
\crefname{lem}{lem.}{lem.s}
\crefname{theorem}{thm.}{thm.s}
\crefname{figure}{fig.}{figs.}
\crefname{table}{tab.}{tabs.}
\crefname{teo}{thm.}{thm.s}
\crefname{assump}{assump.}{assump.s}

\newcommand{\Iverson}[1]{\ensuremath{\llbracket #1 \rrbracket}}

\newcommand{\wrt}{\ensuremath{\text{w.r.t.\ }}}
\newcommand{\st}{\ensuremath{\text{s.t.\ }}}

\newcommand{\citeintext}[1]{\citet{#1}}

\newcommand{\idealpointshort}{\ensuremath{\text{IP}}\xspace}
\newcommand{\roc}{\ensuremath{\text{ROC}}\xspace}
\newcommand{\croc}{\ensuremath{\text{C-ROC}}\xspace}
\newcommand{\iproc}{\ensuremath{\text{\idealpointshort-ROC}}\xspace}
\newcommand{\lambdaroc}{\ensuremath{\lambda\text{-ROC}}\xspace}
\newcommand{\lmcut}{\ensuremath{\text{LMcut}}\xspace}
\newcommand{\iplmcut}{\ensuremath{\text{\idealpointshort-LMcut}}\xspace}
\newcommand{\lambdalmcut}{\ensuremath{\lambda\text{-LMcut}}\xspace}
\newcommand{\cpom}{\ensuremath{\text{C-POM}}\xspace}
\newcommand{\hmax}{\ensuremath{\text{h-max}}\xspace}

\newcommand{\carl}{\ensuremath{\text{CARL}}\xspace}
\newcommand{\astar}{A$^*$\xspace}

\newcommand{\lao}{LAO$^*$\xspace}
\newcommand{\ilao}{i\lao}
\newcommand{\cgilao}{CG-\ilao}
\newcommand{\lrtdp}{LRTDP\xspace}
\newcommand{\idual}{\ensuremath{\text{i-dual}}\xspace}

\newcommand{\iidual}{\ensuremath{\text{i\(^2\)-dual}}\xspace}
\newcommand{\Iidual}{\ensuremath{\text{I\(^2\)-dual}}\xspace}

\newcommand{\lagrange}{\ensuremath{L}\xspace}
\newcommand{\vlambda}{\ensuremath{\vect{\lambda}}\xspace}
\newcommand{\vlambdas}{\ensuremath{\vect{\lambda}\text{s}}\xspace}
\newcommand{\paramssp}[1]{\ensuremath{\ssp(#1)}\xspace}
\newcommand{\lambdassp}{\ensuremath{\paramssp{\vlambda}}\xspace}
\newcommand{\lambdassps}{\ensuremath{\lambdassp\text{s}}\xspace}
\newcommand{\lambdastarssp}{\ensuremath{\paramssp{\vlambda^*}}\xspace}
\newcommand{\vecV}{\ensuremath{\vect{V}}\xspace}
\newcommand{\vecQ}{\ensuremath{\vect{Q}}\xspace}

\newcommand{\vallgreedy}{\ensuremath{\vecV^*_{\forall}}\xspace}

\newcommand{\allgreedy}{\ensuremath{\V^*_{\forall}}\xspace}
\newcommand{\solvelambdassp}{\ensuremath{\textit{solve-}}\ssp}
\newcommand{\findlambdastar}{\ensuremath{\textit{find-}\vlambda^*}}
\newcommand{\findtiedgreedys}{\ensuremath{\textit{find-all-opts-for-\ssp}}}
\newcommand{\extractpi}{\ensuremath{\textit{extract-opt-policy}}}
\newcommand{\subgrad}{\ensuremath{\vect{g}}\xspace}

\newcommand{\lpx}{\ensuremath{x}\xspace}
\newcommand{\lpxs}{\ensuremath{\vect{x}}\xspace}

\newcommand{\flout}{\ensuremath{out}\xspace}
\newcommand{\flin}{\ensuremath{in}\xspace}

\newcommand{\supp}{\ensuremath{\text{supp}}}

\newcommand{\partssp}{\ensuremath{\widehat{\ssp}}\xspace}

\newcommand{\sar}{SAR\xspace}
\newcommand{\elevators}{Elev\xspace}

\newcommand{\exbw}{ExBW\xspace}
\newcommand{\tireworld}{\ensuremath{\text{CTW}}\xspace}
\newcommand{\parc}{PARC\xspace}
\newcommand{\parcn}{PARC\xspace} %

\newcommand{\currp}{\ensuremath{\widehat{\p}_{\V}}\xspace}
\newcommand{\oldp}{\ensuremath{\widehat{\p}_{\text{old}}}\xspace}
\newcommand{\colsToCheck}{\ensuremath{\Gamma}\xspace}
\newcommand{\envelope}{\ensuremath{\mathcal{E}}\xspace}

\newcommand{\residual}{\ensuremath{\textsc{res}}\xspace}

\newcommand{\V}{\ensuremath{V}\xspace}
\newcommand{\Q}{\ensuremath{Q}\xspace}
\newcommand{\Qsa}{\ensuremath{Q(\s, \ac)}\xspace}
\newcommand{\qvalue}{\Q-value\xspace}
\newcommand{\qvalues}{\Q-values\xspace}

\newcommand{\econsistent}{\ensuremath{\epsilon\text{-consistent}}\xspace}
\newcommand{\econsistency}{\ensuremath{\epsilon\text{-consistency}}\xspace}
\newcommand{\strongeconsistent}{strongly \ensuremath{\epsilon\text{-consistent}}\xspace}
\newcommand{\strongeconsistency}{strong \ensuremath{\epsilon\text{-consistency}}\xspace}
\newcommand{\Strongeconsistency}{Strong \ensuremath{\epsilon\text{-consistency}}\xspace}

\begin{document}

\begin{frontmatter}

\paperid{1492}

\title{Solving Constrained Stochastic Shortest Path Problems with Scalarisation}

\author[A,B]{\fnms{Johannes}~\snm{Schmalz}\orcid{0000-0003-3389-2490}\thanks{Corresponding Author. Email: schmalz@cs.uni-saarland.de.}}
\author[B]{\fnms{Felipe}~\snm{Trevizan}\orcid{0000-0001-5095-7132}}

\address[A]{Saarland University, Germany}
\address[B]{Australian National University, Australia}

\begin{abstract}

Constrained Stochastic Shortest Path Problems (CSSPs) model problems with probabilistic effects, where a primary cost is minimised subject to constraints over secondary costs, e.g., minimise time subject to monetary budget.
Current heuristic search algorithms for CSSPs solve a sequence of increasingly larger CSSPs as linear programs until an optimal solution for the original CSSP is found.
In this paper, we introduce a novel algorithm \carl, which solves a series of unconstrained Stochastic Shortest Path Problems (SSPs) with efficient heuristic search algorithms.
These SSP subproblems are constructed with scalarisations that project the CSSP's vector of primary and secondary costs onto a scalar cost. %
\carl finds a maximising scalarisation using an optimisation algorithm similar to the subgradient method which, together with the solution to its associated SSP, yields a set of policies that are combined into an optimal policy for the CSSP.
Our experiments show that \carl solves 50\% more problems than the state-of-the-art on existing benchmarks.

\end{abstract}

\end{frontmatter}

\section{Introduction}

Stochastic Shortest Path Problems (SSPs) model problems where the effects of actions can be probabilistic and Constrained SSPs (CSSPs) extend the model by allowing constraints over secondary costs, enforcing that the incurred secondary costs do not exceed a specified threshold over expectation.
This can represent interesting real-world problems, e.g., planning an aeroplane's route that minimises fuel usage while avoiding bad weather and satisfying certain timing requirements~\cite{Geisser2020:airbus}.
SSPs can be solved with the primal formulation, where we optimise over variables that represent the agent's cost-to-go (a.k.a. value functions); or with the dual formulation over occupation measures, which represent the expected number of times an action is applied in each state.
Algorithms over the primal representation tend to be faster and indeed, the state-of-the-art algorithms for solving SSPs optimally use the primal representation, e.g., \ilao~\cite{Hansen2001:ilao}, \lrtdp~\cite{Bonet2003:lrtdp}, \cgilao~\cite{Schmalz2024:cgilao}.
In contrast, the only heuristic search algorithms for CSSPs use the dual representation: \idual~\cite{trevizan16:idual} and \iidual~\cite{trevizan17:hpom}.
In this paper, we present the first heuristic search algorithm that solves CSSPs optimally using the primal space, and show that it outperforms the existing algorithms.

Our algorithm \carl builds on existing work for finding deterministic policies for CSSPs (which are generally suboptimal)~\cite{Hong2023:DetPiForCSSPs}, with the key difference that \carl finds optimal (potentially stochastic) policies.
\carl works by running heuristic search in the primal space using scalarisation, where a scalarisation \vlambda projects the CSSP's cost vector onto a scalar cost function, thereby inducing an unconstrained SSP.
\carl searches for an optimal scalarisation \(\vlambda^*\) with optimsation techniques similar to the subgradient method~\cite{Shor1985} and solves SSPs induced by the encountered \vlambda scalarisations as subproblems.
Once \carl has found \(\vlambda^*\), it computes the optimal costs-to-go \(\V^*\) associated with that scalarisation, and builds an optimal policy for the CSSP by combining all of \(\V^*\)'s greedy policies.
\carl's subproblems are solved efficiently with heuristic search algorithms for SSPs and its outer problem over scalarisations is blind; this is reversed in the dual methods (\idual and \iidual) where the construction of subproblems is guided with a heuristic, but the subproblems themselves are solved blindly by the LP solver.
In our experiments, this tradeoff pays off and \carl can solve 1264 out of 1290 of our benchmark problems, whereas the state-of-the-art dual methods only solve 808.
For some problems \carl is able to solve all instances while the state-of-the-art is not able to solve any, and over the problems where all algorithms solve all instances \carl offers an average speedup of $10\times$ \wrt the strongest baseline.

\section{Background}\label{sec:background}

\textbf{Stochastic Shortest Path Problems (SSPs)}~\cite{Bertsekas1991:SSPs} are defined by the tuple \sspTuple where \Ss is a finite set of states; \(\sZ \in \Ss\) is the initial state; \(\Sg \subset \Ss\) is a set of goal states that must be reached; \A is a finite set of applicable actions, and we write \(\A(\s)\) to denote the applicable actions in state \s; \pr is the probability transition function where \(\pr(\s'|\s,\ac)\) is the probability of reaching \(\s'\) after applying \ac to \s; and \(\C : \A \to \Reals_{>0}\) is a cost function where \(\C(\ac)\) gives the cost of applying the action \ac.

\textbf{Policies} map states to actions and describe solutions to SSPs.
Policies come in two flavours: \textbf{deterministic policies} \(\p : \Ss \to \A\) and \textbf{stochastic policies} \(\p : \Ss \to \text{distr}(\A)\), which respectively map each state onto a single action, or onto a probability distribution over actions from which an action should be selected randomly;
\(\p(\s,\ac)\) is the probability that \p applies \ac in \s, and we may write \(\p(\s) = \ac\) if \(\p(\s,\ac) = 1\).
The envelope of policy \p from \s, written \(\Ss^{\p, \s}\), denotes the set of states reachable by following \p from \s.
The policy \p is closed \wrt \s if \p is defined for all \(\s' \in \Ss^{\p, \s}\), and \p is proper \wrt \s if following \p from \s guarantees that \Sg is reached with probability 1.
If \p is not closed or not proper it is called open or improper, respectively.
If \s is omitted in these expressions, then we assume \(\s = \sZ\).
We extend the notion of an envelope, and write \(\supp(\p)\) to denote the state-action pairs that might be encountered by \p, i.e., \(\supp(\p) = \{(\s, \ac) : \s \in \Ss^{\p}, \p(\s,\ac) > 0\}\).
We overload the \C symbol and write \(\C(\p)\) to denote the expected cost incurred by following the proper policy \p from \sZ.
We make two standard assumptions for SSPs: a proper policy exists from each state (reachability), and any improper policy incurs infinite cost.
Then, a policy \p is optimal if it is closed and minimises \(\C(\p)\).
Under these assumptions, SSPs always have an optimal deterministic policy~\cite{Bertsekas1991:SSPs}.

\textbf{Primal algorithms for SSPs} work over value functions \(\V : \Ss \to \Reals_{\geq 0}\), which represent the cost-to-go for each state, i.e., \(\V(\s)\) indicates the expected cost that the agent must incur to reach the goal from \s.
For each \V, its \qvalues are \(\Qsa = \C(\ac) + \sum_{\s' \in \Ss} \pr(\s'|\s,\ac) \V(\s')\).
\(\V^*\) is the optimal value function and denotes the cheapest possible cost-to-go for each state.
\(\V^*\) is the unique solution of the Bellman equations:
\[\V(\s) = \min_{\ac \in \A(\s)} \Qsa \; \forall \s \in \Ss \setminus \Sg \text{ and } \V(\s) = 0 \; \forall \s \in \Sg.\]
For \V, its greedy policy \(\p_{\V}\) is defined by \(\p_{\V}(\s) = \text{argmin}_{\ac \in \A(\s)} \Q(\s,\ac)\), where the \(\text{argmin}\) operator breaks ties arbitrarily to yield a single policy.
\(\V^*\)'s greedy policy gives an optimal deterministic policy for the SSP.
To compute \(\V^*\), we can start with some \V and iteratively apply Bellman backups \(\V(\s) \gets \min_{\ac \in \A(\s)} \Qsa\) over states \s.
Value Iteration (VI)~\cite{Bellman57} applies Bellman backups over the whole state space in each step, and converges to \(\V^*\) in the limit as the number of steps increases.

\textbf{Heuristic search algorithms} use heuristic functions to focus on a promising subset of states and actions, and only apply backups there.
For deterministic problems the canonical heuristic-search algorithm is \astar~\cite{Hart1968:astar}, and the state-of-the-art algorithms \ilao~\cite{Hansen2001:ilao}, \lrtdp~\cite{Bonet2003:lrtdp}, and \cgilao~\cite{Schmalz2024:cgilao} generalise \astar to the SSP setting.
A heuristic function is a value function \(\h : \Ss \to \Reals_{\geq 0}\) that estimates \(\V^*\), so that the algorithm can avoid states with large \(\h(\s)\) and prefer states with low \(\h(\s)\).
We require heuristics to be admissible, i.e., to be lower bounds on the optimal value function \(\h(\s) \leq \V^*(\s) \; \forall \s \in \Ss\).
With admissible heuristics the heuristic-search algorithms we consider can guarantee optimality.
For VI, we measure the change in successive value functions with Bellman residual \(\residual(\s) = |\V(\s) - \min_{\ac \in \A(\s)} \Qsa|\), and stop search once the changes in \V are sufficiently small.
In heuristic search, we stop when \V is \econsistent~\cite{Bonet2003:hdp}, i.e., when for some greedy policy \(\p_{\V}\) we have \(\residual(\s) \leq \epsilon ~ \forall \s \in \Ss^{\p_{\V}}\).
This condition guarantees that \(\V = \V^*\) as \(\epsilon \to 0\), and in practice, for sufficiently small \(\epsilon\), \econsistent value functions induce an optimal policy.

\textbf{\cgilao}~\cite{Schmalz2024:cgilao} is a heuristic search algorithm over the primal space, which we explain because it provides crucial features to solve our subproblems efficiently.
\cgilao is presented in \cref{alg:cg-ilao}.
It works over value function \V and a partial SSP \partssp, which contains a subset of the original SSP's states and actions.
Non-goal states in \partssp are called ``expanded'' if they have at least one applicable action in \partssp, and are called ``fringes'' if they have none.
In each step, \cgilao applies Bellman backups over the greedy policy \currp's envelope \envelope, and expands \envelope's fringes.
Thus, \cgilao simultaneously works towards making \currp closed by expanding fringes, and towards making \V \econsistent with the Bellman backups.
Since \partssp need not contain all actions, \cgilao looks for expanded states \s with actions \ac such that  \(\Qsa < \V(\s)\), because this suggests \ac can improve \(\V(\s)\) and should be added if it is missing.
\colsToCheck is a set of all state-action pairs that potentially satisfy \(\Qsa < \V(\s)\), i.e., it is a superset of improving actions.
\colsToCheck is efficiently maintained by tracking changes to \V and \cref{line:cgilao:fixViolatedConstraints} fixes any issues by setting \(\V(\s) \gets \min \{V(\s), \Qsa\}\) for all $(s,a) \in \Gamma$ and adding \ac as required.
Note that this may introduce more instances of \(\V(\s) > \Qsa\), which are themselves recorded in \colsToCheck to be checked later.

\begin{algorithm}[!t]
{\small
\DontPrintSemicolon
  \caption{\cgilao}\label{alg:cg-ilao}

  \function{\cgilao(SSP \ssp, heuristic \(\h\), \(\epsilon \in \Reals_{>0}\))} {
    \(\partssp \gets\) partial SSP containing only \sZ \;
    \(\V \gets \text{value function initialised by } \h\) \;
    \Repeat{\(\text{fringes}(\envelope) = \emptyset\) and \(\oldp = \currp\) and \(\residual \leq \epsilon\)} {
	  \(\currp, \oldp \gets\) greedy policy for \V, restricted to \partssp \;
      \(\envelope \gets\) post-order DFS traversal of \currp from \sZ \;
      \(\partssp, \envelope, \currp \gets\) expand fringes of \envelope \;
      \(\V, \residual, \currp, \colsToCheck \gets\) apply Bellman backups over \envelope, recording increases and decreases to \V in \colsToCheck \;
      \(\V, \residual, \colsToCheck, \partssp \gets\) fix cases of \(\V(\s) > \Qsa\) in  \colsToCheck \; \label{line:cgilao:fixViolatedConstraints}
    } \label{line:cgilao:whileCondition}
    \Return \V
  }
}
\end{algorithm}

\textbf{Constrained SSPs (CSSPs)}~\cite{Altman1995:CMDPs,trevizan16:idual} are extensions to SSPs defined by \(\cssp = \csspTuple\) with two changes from SSPs: (1) the vector cost function \(\vecC : \A \to \Reals_{>0} \times \Reals^{n}_{\geq 0}\), where \(\C[0] : \A \to \Reals_{>0}\) is the primary cost, and \(\C[i] : \A \to \Reals_{\geq 0}\) for \(i \in \{1, \dots, n\}\) are the secondary costs; (2) we have a vector of upper bounds \(\vecUbC \in \Reals^{n}_{\geq 0}\).
Note that we use boldface to denote vectors.
\(\C[i](\p)\) denotes the expected cost of \p on the \(i^{\text{th}}\) cost; we assume reachability and that improper policies incur infinite primary cost.
A policy \p is feasible if it satisfies all its secondary-cost constraints, i.e., \(\C[i](\p) \leq \ubC{i} \; \forall i \in \{1, \dots, n\}\), and \p is optimal if it is closed, feasible, and minimises the primary cost \(\C[0](\p)\) \wrt the other feasible policies.
In contrast to SSPs, CSSPs may have no optimal deterministic policies and only strictly stochastic ones~\cite{Altman1995:CMDPs,trevizan16:idual}.
This difference occurs because there may be deterministic policies that are infeasible on their own, but can be mixed to satisfy the secondary-cost constraints over expectation.

There are no primal method for optimally solving CSSPs and all optimal CSSP algorithms rely on the dual formulation.
The dual formulation optimises over the space of occupation measures \(\lpxs\) where \(\lpx_{\s, \ac}\) represents the expected number of times that \s is reached and then \ac applied.
\ref{lp:om} is the occupation measure LP for CSSPs~\cite{trevizan16:idual}, where \(\Iverson{ \cdot }\) is the Iverson bracket and we use the macros \(\flout(\s) = \sum_{{\ac \in \A(\s)}} \lpx_{\s, \ac}\) and \(\flin(\s) = \sum_{{\s' \in \Ss, \ac' \in \A(\s')}} \lpx_{\s', \ac'} \pr(\s|\s', \ac')\) for each \(\s \in \Ss\).
\ref{lp:om} can be interpreted as a network flow, where a unit of flow is injected into \sZ, and must be routed through the actions so that all of it reaches the goals.
An optimal solution \(\lpxs\) for \ref{lp:om} induces the optimal policy \(\p_{\lpxs}\) with \(\p_{\lpxs}(\s, \ac) = \lpx_{\s, \ac} / \flout(\s) \; \forall \s \in \Ss, \ac \in \A(\s)\).

\newcommand{\lpenvsize}{.85\linewidth}
\resizebox{\lpenvsize}{!}{
\LP{7.5cm}
{\obj{}
{
\min_{\lpxs} \sum_{\mathclap{\s \in \Ss, \ac \in \A(\s)}} \lpx_{\s, \ac} \C[0](\ac) \text{ s.t. \ref{c:om:preservation-of-flow}--\ref{c:om:cssp}}
}
{lp:om} \\
\constr{\flout(\s) - \flin(\s) = \Iverson{ \s = \sZ }}{\forall \s \in \Ss \setminus \Sg}{c:om:preservation-of-flow} \\
\constr{\sum_{g \in \Sg} \flin(g) = 1}{}{c:om:sink} \\
\constr{\lpx_{\s, \ac} \geq 0}{\forall \s \in \Ss, \ac \in \A(\s)}{c:om:non-negativity} \\
\constr{\sum_{\mathclap{\s \in \Ss, \ac \in \A(\s)}} \lpx_{\s, \ac} \C_i(\ac) \leq \ubC{i}}{\forall i \in \{1, \dots, n\}}{c:om:cssp}
}
}
\vspace{0.1cm}

There is an important connection between deterministic and stochastic policies: any stochastic policy can be represented as a convex combination (a.k.a. mixture) of deterministic ones~\cite{Geisser2020:airbus}.
We write this decomposition of the stochastic policy \p as \(\p = \mu_0 \p_0 + \cdots + \mu_k \p_k\) where \(\mu_0, \dots, \mu_k \in \Reals_{> 0}\), \(\mu_0 + \cdots + \mu_k = 1\), and \(\p_0, \dots, \p_k\) are \p's constituent deterministic policies.
We give an example of a CSSP and its policies in \cref{sec:cssp-examples}.

\section{Solving CSSPs with Scalarisation}

In this section we explain how to solve CSSPs with scalarisation, and implement this framework in our novel algorithm \carl.
First, we give a high-level overview of the algorithm, and then give details for its three main steps.

Underlying the scalarisation approach is \ref{lp:vi}, the primal LP for \ref{lp:om}.
Intuitively, the \(\V_{\s}\) variables represent the cost-to-go \(\V(\s)\), so we will use \(\V(\s)\) and \(\V_{\s}\) interchangeably.
In fact, if we construct \ref{lp:vi} for an unconstrained SSP, then the \(\lambda_i\) and \(\ubC{i}\) terms disappear and the LP becomes the standard primal LP for solving SSPs which encodes the Bellman equations for \V.
If we fix \(\vlambda\), then \ref{lp:vi} again encodes the Bellman equations for an SSP, but with the modified cost function \(\C_{\vlambda}(\ac) = [1 \; \vlambda] \cdot \vecC(\ac)\) (from \ref{c:vi:consistency}) and the constant one-time terminal cost \(-\sum_{i=1}^{n} \lambda_i \ubC{i}\) (from the objective).
\lambdassp denotes such an SSP parameterised by \vlambda.
\begin{definition}[Scalarised SSP]\label{def:lambdassp}
Given \cssp and a scalarisation \(\vlambda \in \Reals^{n}_{\geq 0}\), \lambdassp is an SSP relaxation of \cssp with the cost function \(\C_{\vlambda} : \A \to \Reals_{\geq 0}\) s.t. \(\C_{\vlambda}(\ac) = [1 \; \vlambda] \cdot \vecC(\ac)\) and the terminal cost \(-\sum_{i=1}^{n} \lambda_i \ubC{i}\), which is incurred once when a goal is encountered, and is constant for \vlambda.
We write a proper policy's cost as \(\C_{\vlambda}(\p) = [1 \; \vlambda] \cdot \vecC(\p) -\sum_{i=1}^{n} \lambda_i \ubC{i}\).
\end{definition}
We can interpret each \(\vlambda\) as a scalarisation that projects the CSSP's vector cost function onto a scalar cost function.
An optimal solution \(\V^*, \vlambda^*\) for \ref{lp:vi} gives the solution to the scalarised SSP's Bellman equations that are maximal over all scalarisations.
Thanks to the strong duality of LPs, the solution \(\V^*, \vlambda^*\) can be transformed into an optimal solution for \ref{lp:om} using complementary slackness, yielding an optimal stochastic policy for the CSSP.
This transformation can be interpreted as extracting all the greedy deterministic policies for \(\V^*\), and combining them into an optimal policy.

\resizebox{\lpenvsize}{!}{
\LP{7.5cm}{
\obj{}{\max_{\V, \vlambda} \V_{\sZ} - \sum_{i=1}^{n} \lambda_i \ubC{i} \text{ s.t.  \ref{c:vi:consistency}--\ref{c:vi:goal}}}
{lp:vi} \\
\longconstr{
	\V_{\s} \leq \C[0](\ac)
	+ \sum_{i=1}^{n} \lambda_i \C[i](\ac)
	+ \sum_{\mathclap{\s' \in \Ss}}\pr(\s'|\s,\ac)\V_{\s'}
	}{\forall \s \in \Ss\!\setminus\!\St, \ac\!\in\!\A(\s)}{c:vi:consistency}\\
\constr{\V_g = 0}{\forall g \in \St}{c:vi:goal}\\
\constr{\lambda_i \geq 0}{\forall i \in \{1, \dots, n\}}{c:vi:lambda-positivity}
}
}
\vspace{0.2cm}

To avoid solving \ref{lp:vi} directly, we separate the optimisation over \(\V\) and \vlambda into \(\max_{\vlambda} \lagrange(\vlambda)\), where \(\lagrange(\vlambda) = \max_{\V} \V_{\sZ} - \sum_{i=1}^{n} \lambda_i \ubC{i}\), i.e., \(\lagrange(\vlambda)\) is the optimal policy cost for \lambdassp.
If we draw \(\lagrange(\vlambda)\) for all values of \vlambda, we get a surface that is piecewise linear concave~\cite{Hong2023:DetPiForCSSPs}; we give some examples of this in \cref{sec:visualisation-of-L}.
Thus, with an oracle that gives \(\lagrange(\vlambda)\) and a subgradient for each \vlambda, we can use any subgradient method to find \(\vlambda^* = \max_{\vlambda} \lagrange(\vlambda)\).
Finally, we must find \(\V^*\) that describes all optimal deterministic policies for \lambdastarssp, from which we extract an optimal stochastic policy for the CSSP.

\Cref{alg:scalarisation} presents the high-level pseudocode of this scalarisation algorithm, named \carl.
In the remainder of this section, we explain \carl's three steps: (1) finding \(\vlambda^*\) (\cref{line:call-find-lambda-star}), (2) finding \(\V^*\) given \(\vlambda^*\) (\cref{line:call-find-tied-greedys}), and (3) extracting the optimal stochastic policy (\cref{line:call-extract-pi}).
To conclude, we prove \carl's correctness, and give more detail about \carl's error terms.

We point out that \carl follows the concepts of Lagrangian decomposition and uses the same framework for finding \(\vlambda^*\) as the algorithm by \citeintext{Hong2023:DetPiForCSSPs}.
The novelty of \carl is that it produces optimal (potentially stochastic) policies and improves on previous techniques, which we discuss further in \cref{sec:related-work}.

\begin{algorithm}[!t]
{\small
\DontPrintSemicolon
  \caption{Solve CSSP with Scalarisation}\label{alg:scalarisation}

  \function{solve(CSSP \cssp, heuristic \(\vech : \Ss \to \Reals_{\geq 0}^{n+1}\))} {
	\tcp{\(\vlambda^* \in \Reals_{\geq 0}^{n}, \vecV : \Ss \to \Reals_{\geq 0}^{n+1}, \allgreedy : \Ss \to \Reals_{\geq 0}\)}
	\(\vlambda^*, \vecV \gets \findlambdastar(\cssp, \vech)\) \;
	\label{line:call-find-lambda-star}
	\(\allgreedy \gets \findtiedgreedys(\vlambda^*, \vecV, \cssp, \vech)\) \;
	\label{line:call-find-tied-greedys}
	\(\p^* \gets \extractpi(\vlambda^*, \allgreedy, \cssp)\) \;
	\label{line:call-extract-pi}
	\Return{\(\p^*\)}
  }

  \function{\findlambdastar(CSSP \cssp, \(\vech : \Ss \to \Reals_{\geq 0}^{n+1}\))} {
	\label{line:func-find-lambda-star}
	\tcp{\(\vlambda \in \Reals_{\geq 0}^{n}, \vecV : \Ss \to \Reals_{\geq 0}^{n+1}, \subgrad \in \Reals^{n}\)}
	\(\vlambda \gets 0, \vecV \gets \vech\)\;
	\Repeat{\vlambda has converged \textnormal{(see \cref{sec:error-terms})}}{
		\(\vecV \gets \solvelambdassp(\vlambda, \vecV, \cssp)\)\;
		\label{line:call-solve-lambda-ssp}
		subgradient \(\subgrad \gets [\V_1(\sZ) - \ubC{1} \; \cdots \; \V_n(\sZ) - \ubC{n}]\) \;
		\label{line:get-gradient-from-lambda-policy}
		\(\vlambda \gets\) move from \vlambda with \subgrad (see \cref{sec:carl-step-1}) \;
		\label{line:move-with-subgradient}
	}
	\Return{\(\vlambda, \vecV\)} \;
  }

  \function{\(\solvelambdassp(\vlambda, \vecV, \cssp\))} {
	\label{line:func-solve-lambda-ssp}
	\(\vecV \gets \) solve \lambdassp (see \cref{def:lambdassp}) with SSP algorithm modified to work on vector \(\vecV\) (see \cref{sec:carl-step-1}) \;
	\label{line:inner-solve-lambda-ssp}
	\Return{\(\vecV\)}
  }

  \function{\findtiedgreedys(\(\vlambda, \cssp, \vech\))} {
	\label{line:func-find-tied-greedys}
	\(\vallgreedy \gets \solvelambdassp(\vlambda, \vecV, \cssp)\) with modification that satisfies \strongeconsistency (see \cref{eq:extended-econsistency} in \cref{sec:find-vstar-for-lambdastar}) \;
	\label{line:inner-solver-lambda-ssp-with-strong-econsistency}
	\Return{\(\allgreedy \text{ s.t. } \allgreedy(\s) = [1 \; \vlambda^*] \cdot \vallgreedy(\s) \; \forall \s \in \Ss\)}
  }

  \function{\extractpi(\vlambda, \allgreedy, \cssp)} {
	\label{line:func-extract-pi}
	\(\lpxs^* \gets \) solution to \ref{sol:xpi} with \vlambda, \allgreedy, \cssp\;
	\label{line:inner-get-stochastic-pi}
	\Return{\(\p_{\lpxs^*}\)}\;
  }

}
\end{algorithm}

\subsection{Finding a Maximal Scalarisation}\label{sec:carl-step-1}

In this section, we describe \findlambdastar{}~(\cref{line:func-find-lambda-star}), which searches over \vlambdas with a subgradient method (outer problem).
This technique must compute subgradients, which we do by solving \lambdassp as subproblems

\paragraph{Outer optimisation over \vlambda.}
Suppose that we have an oracle that returns \(\lagrange(\vlambda)\) and a subgradient \(\subgrad\) for each \vlambda.
Recall that our optimisation space, i.e., the surface obtained by plotting \(\lagrange(\vlambda)\) for all \vlambdas, is piecewise linear concave.
This function has ``sharp kinks'' and can not be differentiated there, which is why we need subgradients~\cite{Shor1985}.\footnote{Concave functions technically have supergradients, but these are still called subgradients in the literature.}
To find \(\max_{\vlambda} \lagrange(\vlambda)\), we can use any optimisation method that accommodates concave and non-smooth search spaces.
We follow \citeintext{Hong2023:DetPiForCSSPs} and use an exact line search within coordinate search, which has been shown to be efficient for piecewise linear concave models with few constraints.

Coordinate search maximises \(\lagrange(\vlambda)\) by focusing on one coordinate at a time.
That is, it sequentially solves for each \(i \in \{1, \dots, n\}\) the subproblem \(\max_{\lambda_i} \lagrange(\vlambda)\), where all \(\lambda_j\) for \(j \neq i\) are fixed from the previous step.
Thus, for each coordinate \(i\) it maximises \(\lagrange([\lambda_1 \; \dots \; \lambda_n])\) where \(\lambda_i\) is the only variable and all other terms are fixed.
This subproblem is solved with an exact line search which exploits that the subproblem is a piecewise linear concave problem with a single variable.
The approach is reminiscent of binary search: it starts with \(l = 0\) and \(u \in \Reals_{>0}\) which give lower and upper bounds on the optimal assignment to \(\lambda_i\).
Then, the subgradients are computed at \vlambda with \(\lambda_i = l\) and \(\lambda_i = u\), the intersection of these subgradients is computed as \(m\), and either \(l\) or \(u\) is updated to \(m\), depending on \(m\)'s subgradient.
This process repeats until \(l\) and \(u\) converge or their subgradients have the same sign (see \cite{Hong2023:DetPiForCSSPs} for more details).
We visualise how \vlambda is updated with coordinate search in \cref{sec:visualisation-of-L}.
Unfortunately, coordinate search is incomplete for non-smooth problems~\cite{Shi2017:primercoordinatedescentalgorithms}, as we exemplify in \cref{sec:coordinate-search-edge-case}.
Fortunately, these edge cases occur rarely, as we see in our experiments (\cref{sec:experiments}), and are detectable: if we extract a stochastic policy from \(\V^\dagger, \vlambda^\dagger\) whose primary cost is greater than \(\lagrange(\vlambda^\dagger)\), then we know that \(\vlambda^\dagger \neq \vlambda^*\), because \(\lagrange(\vlambda^*) = \C[0](\p^*)\) where \(\p^*\) is extracted from \(\V^*, \vlambda^*\).
If we detect that coordinate search failed we can fall back on a complete method to find \(\vlambda^*\); we use the subgradient method with projection to ensure \(\vlambda \geq 0\)~\cite{Shor1985}.
\Cref{line:move-with-subgradient} represents a single update of \vlambda using coordinate search or the subgradient method as a fall-back, as we have described.

\paragraph{Oracle via \lambdassp Subproblem.}
Recall that \lambdassp (\cref{def:lambdassp}) is an SSP relaxation of the CSSP, and therefore solved optimally by deterministic policies.
\lambdassp's optimal deterministic policy \(\p^*_{\vlambda}\) gives \(\lagrange(\vlambda)\) and a subgradient \(\subgrad \in \Reals^{n}\) in the following way: \(\lagrange(\vlambda) = \C_{\vlambda}(\p^*_{\vlambda})\) and \(\subgrad = [\C[1](\p^*_{\vlambda}) - \ubC{1} \; \cdots \; \C[n](\p^*_{\vlambda}) - \ubC{n}]\)~\cite{Hong2023:DetPiForCSSPs}.
These two steps correspond to lines~\ref{line:call-solve-lambda-ssp} and \ref{line:get-gradient-from-lambda-policy}, respectively.
Thus, we can implement an oracle for \vlambda by finding \(\p^*_{\vlambda}\) and then evaluating it \wrt \(\C_{\vlambda}\) and each secondary-cost function.
To find \(\p^*_{\vlambda}\), we can use any optimal SSP algorithm with any heuristic that is admissible for SSPs, e.g., \hmax~\cite{Bonet2000:hmax}, \lmcut~\cite{Helmert2009:lmcut}, or \roc~\cite{trevizan17:hpom}.
In the remainder of this section, we show how to implement \solvelambdassp{}~(\cref{line:func-solve-lambda-ssp}) efficiently.

\paragraph{Efficient Heuristic Search over \lambdassps.}
The \lambdassp subproblems are unconstrained SSPs and can be solved efficiently with existing heuristic-search algorithms such as \cgilao~\cite{Schmalz2024:cgilao}, but we can improve efficiency further by exploiting some additional properties.
First, instead of using a scalar value function \(\V : \Ss \to \Reals_{\geq 0}\), we use a vector value function \(\vecV : \Ss \to \Reals^{n+1}_{\geq 0}\) with entries for each cost function.
Then, we replace the scalar Bellman backups with a vectorised variant, where a \qvalue is defined as \(\vecQ(\s,\ac) = \vecC(\ac) + \sum_{\s' \in \Ss} \pr(\s'|\s,\ac) \vecV(\s')\) and the Bellman backup becomes \(\vecV(\s) \gets \vecQ(\s,\ac_{\min})\) for some \(\ac_{\min}\) that minimises \([1 \; \vlambda] \cdot \vecQ(\s,\ac)\).
With some minor technicalities (see \cref{sec:vector-bellman-backups} for more details), SSP algorithms can be modified to use this vector Bellman backup, and they still solve \lambdassp as an unconstrained SSP because the vector value function is projected onto \(\V_{\vlambda}(\s) = [1 \; \vlambda] \cdot \vecV(\s)\).
However, we now work on \vecV directly with the advantage that we simultaneously find an optimal policy \(\p^*\) for \lambdassp and evaluate the different costs of \(\p^*\), giving us \(\C[i](\p^*_{\vlambda})\) for each cost function $i$.

To ensure optimality, we require an admissible heuristic to solve \lambdassp, i.e., \(\h(\s) \leq \V_{\vlambda}^*(\s) \; \forall \s \in \Ss\);
in addition, we also require that the heuristic is a vector in $\Reals^{n+1}_{\geq 0}$ so that it can be used with $\vecV$.
To address this, we introduce \(\lambda\) heuristics.
These compute some admissible scalar heuristic \(\h(\s)\), then identify which actions are selected by $\h(s)$ and sum their vector costs to obtain a heuristic cost vector \vech with \([1 \; \vlambda] \cdot \vech(\s) = \h(\s) \leq \V^*_{\vlambda}(\s)\).
For example, if a scalar heuristic uses actions \(\ac_0\) and \(\ac_1\) to construct its estimate \(\h(\s) = [1 \; \vlambda] \cdot \vecC(\ac_0) + [1 \; \vlambda] \cdot \vecC(\ac_1)\), then we assign \(\vech(\s) \gets \vecC(\ac_0) + \vecC(\ac_1)\) which indeed satisfies \([1 \; \vlambda] \cdot \vech(\s) = \h(\s)\).
For most heuristics, we can use their internal data-structures to extract the selected actions, e.g., for \lmcut~\cite{Helmert2009:lmcut} we consider the representative action from each cut and its weight, and for \roc~\cite{trevizan17:hpom} we consider actions' operator counts.
Another way to obtain admissible vector heuristics is with ideal-point (\idealpointshort) heuristics~\cite{Geisser2022:MO_Heuristics}.
Such heuristics are \(\vech = \langle \h_0, \dots, \h_n \rangle\) where \(\h_i\) are all admissible \wrt their cost function \(i\).
The advantage of \(\lambda\) heuristics is that they are computed for the specific \lambdassp and are therefore more informative, but with the trade-off that they have to be recomputed for each \lambdassp, whereas an \idealpointshort heuristic is admissible for all \lambdassps.

\paragraph{Warm starting \lambdassps.}

Notice that all \lambdassps are the same problem only differing in their cost functions.
This makes \vecV from a previous \(\paramssp{\vlambda_\text{old}}\) a good candidate to start solving a new \(\paramssp{\vlambda_\text{new}}\), a technique known as warm start in Operations Research.
However, we have to be careful in reusing \vecV.
Suppose \vecV is a solution to \(\paramssp{\vlambda_{\text{old}}}\) and we have
\(\V_{\vlambda_\text{old}}\) with \(\V_{\vlambda_{\text{old}}}(\s) = [1 \; \vlambda_{\text{old}}] \cdot \vecV(\s) \; \forall \s \in \Ss\) and
\(\V_{\vlambda_\text{new}}\) with \(\V_{\vlambda_{\text{new}}}(\s) = [1 \; \vlambda_{\text{new}}] \cdot \vecV(\s) \; \forall \s \in \Ss\).
Even if \(\V_{\vlambda_{\text{old}}}\) is admissible for \(\paramssp{\vlambda_{\text{old}}}\), there is no guarantee that \(\V_{\vlambda_\text{new}}\) is admissible for \(\paramssp{\vlambda_{\text{new}}}\), which may break the optimality of heuristic search algorithms.
Conveniently, \cgilao has a built-in mechanism for efficiently handling potentially inadmissible states by tracking states whose values have increased or decreased, i.e., we mark the states where \([1 \; \vlambda_{\text{new}}] \cdot \vecV(\s)\) decreases or increases \wrt \([1 \; \vlambda_{\text{old}}] \cdot \vecV(\s)\) in \colsToCheck.
As explained in \cref{sec:background}, this ensures that \cgilao fixes any issues in the value function and finds the optimal policy for \(\paramssp{\vlambda_{\text{new}}}\).
Reusing \vecV in this way speeds up \carl significantly because the optimal policy for one \lambdassp is often good for the next one, and in the best case, if the change in \(\C_{\vlambda}\) does not affect the optimal policy, then \vecV immediately solves the new \(\paramssp{\vlambda'}\).

\subsection{Finding an Optimal Value Function}\label{sec:find-vstar-for-lambdastar}

Once we have \(\vlambda^*\), we need to compute \(\V^* : \Ss \to \Reals_{\geq 0}\) that describes all optimal deterministic policies for \lambdastarssp, which is done in the pseudocode by \findtiedgreedys{} (\cref{line:func-find-tied-greedys}).
Let \(\Pi(\V)\) denote the set of tied-greedy policies for \V, that is, the greedy policies of \V obtained by breaking ties in all possible ways.
Then, if \(\V^*\) is the unique solution to the Bellman equations, \(\Pi(\V^*)\) gives precisely all optimal deterministic policies for the SSP.
For us, \(\V^*\) is produced by an SSP algorithm that only guarantees \econsistency~\cite{Bonet2003:hdp}.
This means \(\V^*\) need only have a small error within one policy's envelope, and any other policies' envelopes can have arbitrarily larger value functions, i.e., \(\V^*\) need only encode a single greedy policy.
We give an example to illustrate this in \cref{sec:strong-econsistency-vs-econsistency}.
In order to obtain a value function that encodes all (approximately) optimal policies, without solving the Bellman equations exactly, we introduce \strongeconsistency.
This condition requires a value function to be \econsistent for all of its greedy policies.
Formally, \V is \strongeconsistent when
\begin{equation}\label{eq:extended-econsistency}
\forall \s \in \bigcup_{\mathclap{\p \in \Pi(\V)}} \Ss^{\p} \quad \residual(\s) \leq \epsilon.
\end{equation}
When a value function is \strongeconsistent, we write it as \allgreedy.
Note that this has not been defined for SSPs before, because there was no need for more than one policy.

\paragraph{Finding \allgreedy.}
Existing SSP algorithms are designed to find \econsistent solutions, so we must modify them to find \strongeconsistent solutions.
This is done by updating the termination condition, and ensuring that Bellman backups are applied for all greedy policies for \V.
We make this modification concrete for the version of \cgilao from \cref{sec:carl-step-1}, but it can be done for other algorithms too.
Recall that this version of \cgilao works over the vector value function \(\vecV : \Ss \to \Reals^{n+1}\), so we are looking for \vallgreedy that induces \allgreedy with \(\allgreedy(\s) = [1 \; \vlambda^*] \cdot \vallgreedy(\s)\).
Instead of tracking the candidate policy \currp, we track the union of all tied-greedy policies (up to \(\epsilon\)) with \(\texttt{tied}(\s) = \{\ac \in \A(\s) : \Q_{\vlambda}(\s, \ac) \leq \min_{\ac' \in \A(\s)} \Q_{\vlambda}(\s, \ac') + \epsilon\}\).
Then, \cgilao's policy envelope \envelope over all tied-greedy policies is found by running the depth-first search from \sZ over \texttt{tied}.
Additionally, any actions with \(\Q_{\vlambda}(\s,\ac) \leq \V_{\vlambda}(\s) + \epsilon\) must be added to \partssp in \cref{line:cgilao:fixViolatedConstraints} so that new tied-greedy actions are caught.
Also, we replace the termination condition \(\oldp = \currp\), now requiring that no new actions have been added to any \(\texttt{tied}(\s)\).
We only care about new actions and do not track removals, because they only shrink the envelope and do not affect \V.
These changes ensure that \cgilao outputs a \strongeconsistent solution which captures all \econsistent policies.
To prove this, we use similar arguments to \citeintext{Schmalz2024:cgilao} to obtain that \(\forall \s \in \envelope \; \residual(\s) \leq \epsilon\).
But \envelope, obtained by DFS over the tied-greedy actions (up to \(\epsilon\)), captures all tied-greedy policies on \V by construction, so \V must be strongly \econsistent.

\subsection{Extracting Stochastic Policies}\label{sec:extract-stoch-policy}

Suppose we have \(\vlambda^*\) and \allgreedy which encodes all optimal deterministic policies for \lambdastarssp.
The final step of \carl is to extract an optimal (potentially stochastic) policy, which is done by \extractpi{} (\cref{line:func-extract-pi}).
To implement this, recall that \allgreedy and \(\vlambda^*\) give the primary cost of all (potentially stochastic) optimal policies for the CSSP with \(\C[0](\p^*) = \allgreedy(\sZ) - \sum_{i=1}^{n} \lambda^*_i \ubC{i}\).
Furthermore, we can extract \(\p^*\)'s envelope and support from the envelopes and supports of \allgreedy's greedy policies.
We now show how to use these insights to extract \(\p^*\), and thereby solve the CSSP optimally.

Given a policy \p, we can evaluate its primary cost \(\C[0](\p)\) with \ref{lp:om} by restricting the LP's variables to the support of \p, adding the constraints \(\lpx_{\s, \ac} / \flout(\s) = \p(\s, \ac)\) for all \(\s, \ac \in \supp(\p)\), and removing the objective function.
The resulting System of Linear Equations and Inequalities (SOL) finds \(\C[0](\p)\) as \(\sum_{\s, \in \Ss, \ac \in \A} \lpx_{\s, \ac} \C[0](\ac)\).
Our case is the converse since we have \p's cost \(\C[0](\p)\) but do not have the distributions \(\p(\s, \cdot)\).
So, instead of adding the constraints \(\lpx_{\s, \ac} / \flout(\s) = \p(\s, \ac)\), we add the constraint \(\sum_{{\s \in \Ss, \ac \in \A(\s)}} \lpx_{\s, \ac} \C[0](\ac) = \C[0](\p^*) = \allgreedy(\sZ) - \sum_{i=1}^{n} \lambda^*_i \ubC{i}\); and we approximate \(\supp(\p^*)\) with \(\supp(\allgreedy)\), which is defined as \(\bigcup_{\p \in \Pi(\allgreedy)} \supp(\p)\), noting that \(\supp(\allgreedy) \supseteq \supp(\p^*)\) as we explain later.
This yields \ref{sol:xpi} without \ref{c:xpi:secondary-lambda-eq-zero} and \ref{c:xpi:secondary-lambda-neq-zero} which we also explain later.
Thus, a solution \(\lpxs\) for \ref{sol:xpi} (with or without \ref{c:xpi:secondary-lambda-eq-zero} and \ref{c:xpi:secondary-lambda-neq-zero}) induces a feasible policy \(\p_{\lpxs}\) with the same cost as an optimal policy by construction.
\ref{sol:xpi}'s only decision is how to distribute the actions in \(\supp(\allgreedy)\) so that the resulting policy has the required costs; in other words, it computes how to combine \lambdassp's deterministic policies into an optimal feasible policy for \cssp.

\resizebox{\lpenvsize}{!}{
\LP{8.5cm}
{\solobj{}
{
\text{find \(\lpxs\) s.t. \ref{c:om:preservation-of-flow}--\ref{c:om:non-negativity} over \(\supp(\allgreedy)\) and \ref{c:xpi:primary-cost}--\ref{c:xpi:secondary-lambda-neq-zero}}
}
{sol:xpi} \\
\constr{\sum_{\mathclap{\s, \ac \in \supp(\allgreedy)}} \lpx_{\s, \ac} \C[0](\ac) = \allgreedy(\sZ) - \sum_{i=1}^{n} \lambda^*_i \ubC{i}}{}{c:xpi:primary-cost} \\
\constr{\sum_{\mathclap{\s, \ac \in \supp(\allgreedy)}} \lpx_{\s, \ac} \C_i(\ac) \leq \ubC{i}}{\forall i \in \{1, \dots, n\} \text{ s.t. } \lambda^*_i = 0}{c:xpi:secondary-lambda-eq-zero} \\
\constr{\sum_{\mathclap{\s, \ac \in \supp(\allgreedy)}} \lpx_{\s, \ac} \C_i(\ac) = \ubC{i}}{\forall i \in \{1, \dots, n\} \text{ s.t. } \lambda^*_i > 0}{c:xpi:secondary-lambda-neq-zero}
}
}
\vspace{0.1cm}

Formally, \ref{sol:xpi} comes from complementary slackness, a technique that lets us transform an optimal solution of an LP into an optimal solution for its dual~\cite{Bertsimas1997:LPs}.
Suppose \(\allgreedy, \vlambda^*\) is an optimal basic feasible solution for \ref{lp:vi} that is non-degenerate; intuitively, a basic feasible solution corresponds to a deterministic policy (rather than a stochastic one) and non-degeneracy is a technical requirement that the solution does not have too many variables set to zero.
With such \(\allgreedy, \vlambda^*\), we can recover a solution for \ref{lp:vi}'s dual (\ref{lp:om}) with complementary slackness by constructing a system of linear equations according to the following rules, and then solving it:

\resizebox{\lpenvsize}{!}{
\begin{minipage}{\linewidth}
\[
\begin{aligned}
\allgreedy(\s) > 0 \implies &\text{ \ref{c:om:preservation-of-flow} for \s is tight, i.e., } \flout(\s) - \flin(\s) = \Iverson{ \s = \sZ } \\
\lambda_i > 0 \implies &\text{ \ref{c:om:cssp} for \(i\) is tight, i.e., } \sum_{\mathclap{\s \in \Ss, \ac \in \A(\s)}} \lpx_{\s, \ac} \C_i(\ac) = \ubC{i} \\
\end{aligned}
\]
\[
\begin{aligned}
\text{\ref{c:vi:consistency} for \s, \ac is loose, i.e., } \allgreedy(\s) < \Q^*_{\forall}(\s,\ac)  \implies \lpx_{\s, \ac} = 0.
\end{aligned}
\]
\end{minipage}
}
\vspace{0.1cm}

These rules describe the constraints of \ref{sol:xpi}.
In particular, note that these rules tighten the secondary-cost constraints to an equality
\(\sum_{{\s \in \Ss, \ac \in \A(\s)}} \lpx_{\s, \ac} \C_i(\ac) = \ubC{i}\) whenever \(\lambda^*_i > 0\), yielding the constraints \ref{c:xpi:secondary-lambda-eq-zero} and \ref{c:xpi:secondary-lambda-neq-zero}.
Unfortunately, our \(\allgreedy, \vlambda^*\) is often degenerate, so this system of linear equations is underspecified, and we must include additional constraints from the original problem.
In particular, we have to reintroduce the secondary cost constraints \ref{c:om:cssp} where \(\lambda_i = 0\), which is what we do in \ref{c:xpi:secondary-lambda-eq-zero}.
We also include \ref{c:xpi:primary-cost} to ensure the resulting policy is indeed optimal.

We have been claiming that \(\supp(\p^*) \subseteq \supp(\allgreedy)\) and that \ref{sol:xpi} is constructing \(\p^*\) from the optimal deterministic policies of \lambdastarssp.
This follows from the last complementary slackness rule: occupation measures \(\lpx_{\s,\ac}\) are only allowed to be non-zero if \(\allgreedy(\s) = \Q^*_{\forall}(\s,\ac)\), i.e., if \ac is a tied-greedy action and thereby \(\p(\s) = \ac\) for some \(\p \in \Pi(\allgreedy)\).

\subsection{Correctness of \carl}\label{sec:theory}

In this section we show that \carl always finds an optimal solution.
First, \(\findlambdastar(\cssp, \vech)\)~(\cref{alg:scalarisation}) is guaranteed to find \(\vlambda^*\).
This is because \(\max_{\vlambda} \lagrange(\vlambda)\) is a piecewise linear concave problem, solving \lambdassps yields subgradients~\cite{Hong2023:DetPiForCSSPs}, and we use a complete subgradient method that is guaranteed to converge to \(\vlambda^*\) (if coordinate search fails, it falls back on the subgradient method, see \cref{sec:carl-step-1}).
It remains to show that \carl finds \(\V^*\) for \lambdastarssp such that the solution encodes an optimal policy.

\begin{lemma}\label{lem:optimal-pi-has-optimal-constituent-policies}
If \(\p^*\) is an optimal policy for the CSSP, then its constituent deterministic policies must be optimal for \(\lambdastarssp\).
\end{lemma}

\begin{proof}
We first show that \(\p^*\) is optimal for \lambdastarssp, and then conclude that its constituent deterministic policies must also be optimal for \lambdastarssp.
Policy \(\p^*\)'s cost in \lambdastarssp can be rewritten as \(\C_{\vlambda^*}(\p^*) = \C[0](\p^*) + \sum_{i=1}^{n} \lambda^*_i (\C[i](\p^*) - \ubC{i})\).
But \(\p^*\) is feasible, so \(\C[i](\p^*) - \ubC{i} \leq 0\) for all \(i \in \{1, \dots, n\}\), and therefore \(\C_{\vlambda^*}(\p^*) \leq \C[0](\p^*)\).
On the other hand, we know that \(\C_{\vlambda^*}(\p) \geq \C[0](\p^*)\) for any policy \p, due to the semantic of \(\vlambda^*\) within strong duality.
Thus, \(\C_{\vlambda^*}(\p^*) = \C[0](\p^*)\), which is the minimal cost for \lambdastarssp, and \(\p^*\) is optimal for \lambdastarssp.
We know that \(\p^*\) can be decomposed into the convex combination of deterministic policies \(\mu_{0} \p_{0} + \cdots + \mu_{k} \p_{k}\), and \(\C_{\vlambda^*}(\p_i) = \C_{\vlambda^*}(\p^*)\) for each \(i \in \{0, \dots, k\}\), because otherwise \(\p^*\) can not be optimal for \lambdastarssp.
\end{proof}

\begin{theorem}\label{thm:findTiedGreedysIsCorrect}
\carl's \(\findtiedgreedys(\vlambda^*, \cssp, \vech)\) finds \vallgreedy so that \ref{sol:xpi} with \(\vlambda^*\) and \allgreedy constructed by \(\allgreedy(\s) = \vlambda^* \cdot \vallgreedy(\s)\) produces an optimal stochastic policy.
\end{theorem}

\begin{proof}
Suppose \allgreedy encodes all the optimal deterministic policies for \lambdastarssp.
If an optimal policy \(\p^*\) exists for the CSSP, then its constituent deterministic policies must be optimal for \lambdastarssp by \cref{lem:optimal-pi-has-optimal-constituent-policies}; therefore its constituent policies are a subset of \(\Pi(\allgreedy)\).
It follows that \ref{sol:xpi} contains the whole support of such policy \(\p^*\).
Since \(\C[0](\p^*) = \vallgreedy(\sZ) - \sum_{i=1}^{n} \lambda_i^* \ubC{i}\) due to strong duality, it must be the case that \ref{sol:xpi} induces an optimal policy, thanks to its connection to \ref{lp:om}.
We have argued that the version of \cgilao presented in \cref{sec:find-vstar-for-lambdastar} finds all policies for which \V can be \econsistent, which gives approximately optimal policies for small \(\epsilon\), and precisely the optimal policies as \(\epsilon \to 0\).
We give more detail about \carl's error terms in \cref{sec:error-terms}.
\end{proof}

\subsection{Error Terms}\label{sec:error-terms}

In this section, we describe the error terms that are present in \carl and explain how to bound their effects on \carl's policies.
We have already discussed that \cgilao, which solves \carl's \lambdassp subproblems, uses \strongeconsistency with \(\epsilon \in \Reals_{>0}\) as its stopping condition, extending regular \econsistency for SSPs (see \cref{sec:find-vstar-for-lambdastar}).
The error associated with \econsistency is well-understood and accepted in the SSP literature: for a fixed \(\epsilon\) it is possible to construct a pathological problem where an \econsistent policy is arbitrarily worse than the optimal policy, but for any SSP, as \(\epsilon \to 0\), \econsistent policies become optimal~\cite{Bonet2003:hdp,Mausam2012:MDPs,Hansen2015:UB-for-V}.
We inherit the same properties in \strongeconsistency, so our version of \cgilao produces \(\V'_{\forall} \approx \allgreedy\) where the error disappears as \(\epsilon \to 0\).
Another error term appears in \findlambdastar{}, which we have omitted until now for simplicity.
Recall that to find \(\vlambda^*\) we use coordinate search, and if it fails we fall back on the subgradient method (see \cref{sec:carl-step-1}).
Both require a tolerance term \(\eta \in \Reals_{>0}\).
In practice, coordinate search stops when it iterates over all coordinates and no single coordinate was able to improve \(\lagrange(\vlambda)\) by more than \(\eta\).
The subgradient method stops when its step size (which is monotonically decreasing over the algorithm's iterations) is smaller than \(\eta\).
In both cases, this yields an approximate solution \(\vlambda' \approx \vlambda^*\), where the error disappears as \(\eta \to 0\).

Both \(\epsilon\) and \(\eta\) can affect \carl's solution, and we explain how this can be mitigated.
Assume for now that \(\epsilon\) is sufficiently small.
If \carl finds \(\vlambda' \approx \vlambda^*\) and an associated policy \p, we have
\[\lagrange(\vlambda') \leq \C[0](\p^*) \leq \C[0](\p)\]
where \(\p^*\) is an optimal policy for the CSSP.
This gives us the optimality gap \(\C[0](\p) - \lagrange(\vlambda')\), which upper bounds how far \p's primary cost deviates from \(\p^*\)'s.
If the gap is too large, we can reduce \(\eta\).
Now, we return our attention to \(\epsilon\).
The value of \(\epsilon\) affects the subgradient search oracle's value of \(\lagrange(\vlambda)\) and subgradient \subgrad at each \vlambda, which in turn can significantly affect the found policy's quality and breaks the optimality gap for \(\eta\).
Recall that this is standard for algorithms using \econsistency, and they only guarantee an optimal policy as \(\epsilon \to 0\).
For additional guarantees, there are ``outer loop'' methods that run the algorithm and select \(\epsilon\) automatically in order to give the stronger guarantee of \(\epsilon\)-optimality~\cite{Hansen2015:UB-for-V,Hansen2017}.
\carl is in a similar position, and guarantees an optimal policy as \(\epsilon \to 0\), but requires an outer loop method to select \(\epsilon\) automatically for stronger guarantees.
The outer loop methods for ensuring \(\epsilon\)-optimality apply to the strongly \econsistent subproblems with minimal changes, which lets us recover a correct optimality gap.
Thus, it is possible to modify \carl to produce solutions with stronger guarantees.
In practice, values of \(\epsilon = \eta = 0.0001\) are usually ``good enough.''

\section{Related Work}\label{sec:related-work}

In the Operations Research (OR) community, it is common to consider the Lagrangian dual of a constrained problem, and then use some version of the subgradient method to find a solution.
This framework has been used to find plans for constrained deterministic shortest path problems~\cite{Handler1980:ConstrainedShortestPath}, stochastic policies for Constrained Partially Observable Markov Decision Processes (POMDPs)~\cite{Lee2018:ConstrainedPOMDP}, and deterministic policies for CSSPs~\cite{Hong2023:DetPiForCSSPs}.
Our presentation differs from theirs, but \ref{lp:vi} is functionally identical to the Lagrangian dual of \ref{lp:om}, and \carl fits within this OR framework.
They have the same \vlambda as our scalarisation term, but they call it a Lagrangian multiplier.

\citeintext{Hong2023:DetPiForCSSPs}'s algorithm is the most similar to \carl.
Both solve CSSPs and start by finding the CSSPs' \(\vlambda^*\).
The difference is that \carl looks for optimal, potentially stochastic, policies whereas they focus on deterministic ones.
Our method for finding \(\vlambda^*\) is based on theirs but we make substantial algorithmic improvements.
They solve \lambdassps with \ilao and reuse \vecV as a warm start (see \cref{sec:carl-step-1}); however, they need to run VI on \ilao's expanded states to make \vecV admissible for the next step.
\carl uses \cgilao, allowing it to focus only on the changes to \(\vlambda\) as opposed to VI's blind search.
As a result, \carl finds \(\vlambda^*\) up to \(4.7\times\) faster than their method with a mean speedup of \(2\times\).

Once \(\vlambda^*\) has been found, the similarities between our algorithms end.
Strong duality does not hold for deterministic policies, thus \(\vlambda^*\) does not directly produce an optimal deterministic policy for the original CSSP.
To overcome this, they use an expensive second stage that enumerates deterministic policies until the optimal one is found.
\carl does not require this stage because strong duality holds for stochastic policies.
Instead, we must ensure that \(\V^*\) encodes all optimal policies for \lambdastarssp, which can be done cheaply with dynamic programming, and then we use \ref{sol:xpi} to extract the optimal policy for the CSSP.
Both these steps have no analogue in their algorithm.
Thus, \carl is guaranteed to find an optimal policy, stochastic or not, for CSSPs with relatively little computational effort after finding \(\vlambda^*\), whereas \citeintext{Hong2023:DetPiForCSSPs}'s method can only find an optimal \textit{deterministic} policy, which can cost more than the optimal stochastic policy, and requires an expensive second stage to do so.

The algorithm by \citeintext{Lee2018:ConstrainedPOMDP} shares with \carl that they get deterministic policies from their subproblems, which must be combined into a stochastic one.
However, their setting and approach is quite different: they use Monte-Carlo sampling to approximate their \qvalues, and their policies map histories onto action distributions, obtained by solving an LP for each history.
We compute \(\V^*\) and its \qvalues with a modified SSP search, and then solve \ref{sol:xpi} once.

Multi-Objective (MO) planning is also relevant.
There are various models of probabilistic MO problems, e.g., MOMDPs with queries from model checking~\cite{Forejt2012:scalarisation} and MOSSPs from planning~\cite{Chen2023:MOSSP}.
These models are similar to CSSPs in that they have multiple cost functions, but the difference is that MO problems are solved by finding all undominated trade-offs between cost functions, e.g.,  consider \(\p_1, \p_2, \p_3\) with \(\vecC(\p_1) = [1 \; 3], \vecC(\p_2) = [3 \; 1], \vecC(\p_3) = [3 \; 2]\); \(\p_1\) and \(\p_2\) are undominated and give different trade-offs of \(\C_0\) and \(\C_1\), and \(\p_3\) is dominated by \(\p_2\).
The set of undominated policies is called the Pareto front.
In contrast, CSSPs are solved by a single policy that minimises \(\C_0\) while satisfying the secondary-cost constraints.
Consequently, MOSSPs are more general than CSSPs, and their algorithms can be applied to solve CSSPs, e.g., \cite{Forejt2012:scalarisation,Chen2023:MOSSP}, but these are much less efficient because they need to track an entire set of policies in the Pareto front, rather than searching for a single policy.

Scalarisation is a standard technique in MO planning~\cite{Ehrgott2006:MO}, and has been applied to MOMDPs with queries~\cite{Forejt2012:scalarisation}.
Their algorithm's similarity to \carl is that both project the cost vector onto scalars to induce single-objective SSPs (or MDPs) as their subproblems.
However, the way that the scalarisations are explored and the purpose is completely different: they use different scalarisations to find different trade-offs, and \carl uses scalarisation in the context of Lagrangian optimisation where scalarisations \vlambda are also called Lagrangian multipliers.
Importantly, \carl's scalarisation is a single-objective technique, and it progressively finds a better scalarisation vector, whereas \citeintext{Forejt2012:scalarisation} accumulate a set of non-dominated scalarisation vectors.
Another difference is that \citeintext{Forejt2012:scalarisation} use Value Iteration to solve the subproblems, and we use heuristic search, which increases \carl's efficiency but presents technical challenges as addressed in \cref{sec:carl-step-1} and \cref{sec:find-vstar-for-lambdastar}.
Heuristic search has been shown to dominate non-guided search approaches from model checking, e.g., \cite{baumgartner18:pltldual}.
We point out that our method for efficiently solving subproblems can be applied to the algorithm from \citeintext{Forejt2012:scalarisation}, establishing another contribution of this work.

\section{Experiments}\label{sec:experiments}

We compare our novel algorithm \carl with \idual~\cite{trevizan16:idual} and \iidual~\cite{trevizan17:hpom}, the state-of-the-art algorithms for finding optimal stochastic policies for CSSPs.
For \carl, we consider the heuristics \lmcut~\cite{Helmert2009:lmcut} and \roc~\cite{trevizan17:hpom} as \(\lambda\) heuristics (\cref{sec:carl-step-1}), called \lambdalmcut and \lambdaroc.
In \cref{sec:more-results} we also test \lmcut and \roc as ideal-point (\idealpointshort) heuristics (\cref{sec:carl-step-1}).
For \idual, we use ideal-point \lmcut (\iplmcut) and \croc~\cite{trevizan17:hpom}.
I\(^2\)-dual uses \cpom~\cite{trevizan17:hpom}, which is built into the algorithm.
In our test problems, we allow CSSPs to have dead ends, but transform them into CSSPs without dead ends with the finite-penalty transformation, which adds give-up actions to each state that lead to a goal with a large cost penalty.
We require a user-specified dead-end penalty for each cost function~\cite{trevizan16:idual}.
For problems with dead ends before the finite-penalty transformation, we augment \roc and \iidual's built-in \cpom with \hmax as a dead-end detector~\cite{trevizan17:hpom}.
We run experiments over the following benchmark domains:

\paragraph{Search and Rescue~(\sar)~\cite{trevizan16:idual}.}
A drone must rescue one of multiple survivors on an \(n \times n\) grid as quickly as possible (\(n \in \Naturals\)).
One survivor's position is known and has distance \(d \in \Naturals\) from the drone's initial position.
Some other locations, selected according to a density \(r \in [0, 1]\), have a predetermined probability of having another survivor.
There is a single secondary-cost constraint over fuel usage.

\paragraph{Elevators~(\elevators)~\cite{trevizan16:idual}.}
In a 10-floor building, \(w \in \Naturals\) people have called the elevator and are waiting, and \(h \in \Naturals\) hidden people have not pressed the button yet, and do so probabilistically in each timestep.
The \(e \in \Naturals\) elevators must be routed to deliver all people to their destinations in the fewest timesteps.
Each person has a constraint over the time spent waiting, and the time spent travelling in an elevator, resulting in \(2(w+h)\) secondary-cost constraints.

\paragraph{Exploding Blocks World~(\exbw)~\cite{trevizan16:idual}.}
Based on Exploding Blocks World for SSPs~\cite{Bonet2006:IPPC}, we must rearrange \(N \in \Naturals\) blocks into a given arrangement with the minimal number of moves.
This is complicated by the blocks being rigged with an explosive that can explode, destroying the block or table immediately underneath it.
In the CSSP variant, the table can be repaired with a large primary cost, but destroyed blocks can not be repaired.
There is a single secondary-cost constraint, limiting the expected number of destroyed blocks to \(c \in \Reals_{\geq 0}\).
We consider specific starting and goal arrangements of blocks from the 2008 International Probabilistic Planning Competition~\cite{Bryce2008:ippc}, and identify specific problems by the parameter \(id\).

\paragraph{PARC Printer~(\parc)~\cite{trevizan17:hpom}.}
Based on the IPC domain, a modular printer must print four pages with various requirements.
Three of the printer's components are unreliable, where a page can get jammed with probability 0.1, simultaneously ruining the page and rendering the component unusable.
There are two secondary-cost constraints: \(f \in [0, 1]\) constrains how many components are allowed to fail and \(u \in \Naturals\) limits how often a particular component is allowed to be used (\(u = \infty\) deactivates this constraint).

\paragraph{Triangle Tireworld~(\tireworld)~\cite{Schmalz2024:det-pi-for-cssp}.}
In the Triangle Tireworld for SSPs we must drive across a triangular network of cities, where \(n \in \Naturals\) specifies the size of the network, and \(d \in \Naturals\) gives the starting distance from the goal~\cite{Little2007:PlannerVsReplanner,Schmalz2024:cgilao}.
Between any two cities the car gets a flat tyre with probability \(0.5\); with a spare tyre the flat can be replaced and the journey can continue; if no spare is loaded, the car is stuck and no actions are available.
The car only fits one spare tyre at a time, and spare tyres can only be acquired in select cities.
In the CSSP variant, tyres can be purchased in these cities for one of \(c \in \Naturals\) currencies, and the problem has secondary-cost constraints over the \(c\)-many currencies.

We ran each planner and heuristic on 30 instances of each problem on a cluster with Intel Xeon 3.2 GHz CPUs limited to one CPU core with 30 mins and 4GB RAM.
For \sar and \elevators, a problem instance is a randomly generated problem for the specified parameters.
For the other domains,
the parameters specify a unique problem and an instance couples this problem with a random seed that is passed to the algorithm.
We use CPLEX Version 22.1.1 to solve LPs and SOLs.
The source code and benchmarks are available at \cite{Schmalz2025:CARLSourceCode}.

\begin{table*}[ht!]
\resizebox{1.01\textwidth}{!}{
\setlength\tabcolsep{2px}

\begin{tabular}[t]{|ll| S[table-format=2] S[table-format=4.1(3.1)]| S[table-format=2] S[table-format=4.1(3.1)]| S[table-format=2] S[table-format=4.1(3.1)]| S[table-format=2] S[table-format=4.1(3.1)]| S[table-format=2] S[table-format=4.1(3.1)]|}
\hline
 \multicolumn{2}{|c|}{algorithm} & \multicolumn{4}{c|}{CARL} & \multicolumn{4}{c|}{i-dual} & \multicolumn{2}{c|}{i2-dual} \\
 \multicolumn{2}{|c|}{heuristic} & \multicolumn{2}{c}{\lambdaroc} & \multicolumn{2}{c|}{\lambdalmcut} & \multicolumn{2}{c}{\croc} & \multicolumn{2}{c|}{\iplmcut} & \multicolumn{2}{c|}{} \\
\hline
\multirow[c]{6}{*}{\rotatebox{90}{{\centering \sar (n, d, r)}}} & 4, 3, .75 & \bestPerformerCROC \bestPerformer 30 & \bestPerformerCROC \bestPerformer 5.6(1.4) & \bestPerformerLMCut \bestPerformer 30 & \bestPerformerLMCut \bestPerformer 4.3(1.0) & 30 & 104.1(51.0) & 30 & 45.8(21.1) & 30 & 69.2(56.4) \\
 & 4, 4, .75 & \bestPerformerCROC \bestPerformer 30 & \bestPerformerCROC \bestPerformer 41.9(5.3) & \bestPerformerLMCut \bestPerformer 30 & \bestPerformerLMCut \bestPerformer 34.1(4.2) & 5 & 1345.7(534.1) & 9 & 979.8(296.6) & 8 & 495.5(417.8) \\
 & 5, 3, .50 & \bestPerformerCROC \bestPerformer 30 & \bestPerformerCROC \bestPerformer 4.9(0.9) & \bestPerformerLMCut \bestPerformer 30 & \bestPerformerLMCut \bestPerformer 4.8(1.1) & 30 & 25.1(8.5) & 30 & 13.1(4.6) & 30 & 11.1(4.0) \\
 & 5, 3, .75 & \bestPerformerCROC \bestPerformer 30 & \bestPerformerCROC \bestPerformer 12.3(2.5) & \bestPerformerLMCut \bestPerformer 30 & \bestPerformerLMCut \bestPerformer 13.3(2.9) & 30 & 253.8(110.6) & 30 & 113.3(46.0) & 30 & 168.0(105.9) \\
 & 5, 4, .50 & \bestPerformerCROC \bestPerformer 30 & \bestPerformerCROC \bestPerformer 30.8(6.8) & \bestPerformerLMCut \bestPerformer 30 & \bestPerformerLMCut \bestPerformer 36.2(7.9) & 17 & 695.0(268.5) & 24 & 550.6(166.1) & 27 & 637.8(213.1) \\
 & 5, 4, .75 & \bestPerformerCROC \bestPerformer 30 & \bestPerformerCROC \bestPerformer 126.6(45.2) & \bestPerformerLMCut \bestPerformer 30 & \bestPerformerLMCut \bestPerformer 142.5(49.3) & 6 & 930.9(369.3) & 6 & 499.2(226.4) & 8 & 349.1(288.1) \\
\hline \multirow[c]{7}{*}{\rotatebox{90}{{\centering \elevators (e,w,h)}}} & 1, 1, 1 & \bestPerformerCROC 30 & \bestPerformerCROC 1.6(0.6) & 30 & 1.2(0.3) & \bestPerformerCROC 30 & \bestPerformerCROC 1.5(0.1) & \bestPerformerLMCut \bestPerformer 30 & \bestPerformerLMCut \bestPerformer 0.8(0.1) & 30 & 6.2(0.6) \\
 & 1, 1, 2 & 29 & 50.9(26.8) & 29 & 35.0(12.7) & \bestPerformerCROC 30 & \bestPerformerCROC 25.3(1.0) & \bestPerformerLMCut \bestPerformer 30 & \bestPerformerLMCut \bestPerformer 21.1(2.6) & 30 & 280.0(46.5) \\
 & 1, 2, 1 & 25 & 14.2(5.6) & 25 & 13.1(4.2) & \bestPerformerCROC 30 & \bestPerformerCROC 13.9(0.9) & \bestPerformerLMCut \bestPerformer 30 & \bestPerformerLMCut \bestPerformer 9.8(1.4) & 30 & 174.2(38.4) \\
 & 1, 2, 2 & 22 & 313.3(150.2) & 22 & 336.9(115.8) & \bestPerformerCROC \bestPerformer 30 & \bestPerformerCROC \bestPerformer 340.1(35.7) & \bestPerformerLMCut \bestPerformer 30 & \bestPerformerLMCut \bestPerformer 392.5(72.1) & 1 & 989.2 \\
 & 2, 1, 1 & \bestPerformerCROC \bestPerformer 30 & \bestPerformerCROC \bestPerformer 104.5(23.2) & \bestPerformerLMCut \bestPerformer 30 & \bestPerformerLMCut \bestPerformer 121.8(26.7) & 23 & 670.6(173.7) & 26 & 576.1(181.4) & 1 & 1799.7 \\
 & 2, 1, 2 & \bestPerformerCROC \bestPerformer 20 & \bestPerformerCROC \bestPerformer 1321.5(161.1) & \bestPerformerLMCut 3 & \bestPerformerLMCut 1344.2(341.5) & 1 & 1799.6 & 1 & 1799.1 & 0 &  \\
 & 2, 2, 1 & \bestPerformerCROC \bestPerformer 28 & \bestPerformerCROC \bestPerformer 569.2(131.7) & \bestPerformerLMCut 23 & \bestPerformerLMCut 890.5(164.2) & 5 & 1608.9(229.1) & 5 & 1444.4(352.4) & 0 &  \\
\hline \multirow[c]{9}{*}{\rotatebox{90}{{\centering \exbw (id,N,c)}}} & 1, 5, .1 & \bestPerformerCROC 30 & \bestPerformerCROC 8.6(0.1) & \bestPerformerLMCut \bestPerformer 30 & \bestPerformerLMCut \bestPerformer 1.6(0.0) & 30 & 97.2(1.0) & 30 & 7.9(0.1) & 30 & 626.6(18.8) \\
 & 2, 5, .07 & \bestPerformerCROC 30 & \bestPerformerCROC 40.4(0.3) & \bestPerformerLMCut \bestPerformer 30 & \bestPerformerLMCut \bestPerformer 18.7(0.1) & 0 &  & 30 & 732.9(30.6) & 0 &  \\
 & 3, 6, .91 & \bestPerformerCROC 30 & \bestPerformerCROC 173.5(5.1) & \bestPerformerLMCut \bestPerformer 30 & \bestPerformerLMCut \bestPerformer 7.5(0.1) & 0 &  & 30 & 10.9(0.2) & 0 &  \\
 & 4, 6, .16 & \bestPerformerCROC 30 & \bestPerformerCROC 161.2(1.4) & \bestPerformerLMCut \bestPerformer 30 & \bestPerformerLMCut \bestPerformer 63.7(0.5) & 0 &  & 27 & 1462.2(79.6) & 0 &  \\
 & 5, 7, .01 & \bestPerformerCROC 30 & \bestPerformerCROC 30.7(0.2) & \bestPerformerLMCut \bestPerformer 30 & \bestPerformerLMCut \bestPerformer 25.8(0.2) & 30 & 82.7(1.5) & 30 & 38.5(0.6) & 30 & 451.0(14.5) \\
 & 6, 8, .3 & \bestPerformerCROC 30 & \bestPerformerCROC 300.4(10.9) & \bestPerformerLMCut \bestPerformer 30 & \bestPerformerLMCut \bestPerformer 91.6(0.6) & 0 &  & 0 &  & 0 &  \\
 & 7, 8, .5 & \bestPerformerCROC 30 & \bestPerformerCROC 0.3(0.0) & \bestPerformerLMCut \bestPerformer 30 & \bestPerformerLMCut \bestPerformer 0.2(0.0) & 30 & 2.2(0.1) & 30 & 0.9(0.1) & 30 & 12.1(0.4) \\
 & 8, 8, .63 & \bestPerformerCROC \bestPerformer 30 & \bestPerformerCROC \bestPerformer 35.1(0.5) & \bestPerformerLMCut \bestPerformer 30 & \bestPerformerLMCut \bestPerformer 34.8(0.4) & 6 & 1799.2(0.1) & 13 & 1691.9(45.5) & 0 &  \\
 & 9, 8, .4 & \bestPerformerCROC 30 & \bestPerformerCROC 177.9(1.4) & \bestPerformerLMCut \bestPerformer 30 & \bestPerformerLMCut \bestPerformer 14.9(0.1) & 0 &  & 30 & 202.3(7.5) & 0 &  \\
\hline \end{tabular}

~
\hspace{-1.9mm}

\begin{tabular}[t]{|ll| S[table-format=2] S[table-format=3.1(2.1)]| S[table-format=2] S[table-format=3.1(2.1)]| S[table-format=2] S[table-format=4.1(2.1)]| S[table-format=2] S[table-format=4.1(2.1)]| S[table-format=2] S[table-format=4.1(3.1)]|}
\hline
 \multicolumn{2}{|c|}{algorithm} & \multicolumn{4}{c|}{CARL} & \multicolumn{4}{c|}{i-dual} & \multicolumn{2}{c|}{i2-dual} \\
 \multicolumn{2}{|c|}{heuristic} & \multicolumn{2}{c}{\lambdaroc} & \multicolumn{2}{c|}{\lambdalmcut} & \multicolumn{2}{c}{\croc} & \multicolumn{2}{c|}{\iplmcut} & \multicolumn{2}{c|}{} \\
\hline \multirow[c]{12}{*}{\rotatebox{90}{{\centering \parcn (f,u)}}} & 0.0, 1 & 30 & 311.2(13.6) & 30 & 270.9(8.3) & \bestPerformerCROC 30 & \bestPerformerCROC 126.2(9.1) & \bestPerformerLMCut 30 & \bestPerformerLMCut 103.1(2.4) & \bestPerformer 30 & \bestPerformer 42.8(3.1) \\
 & 0.0, \(\infty\) & 30 & 100.7(0.8) & 30 & 141.3(1.0) & \bestPerformerCROC 30 & \bestPerformerCROC 47.1(2.5) & \bestPerformerLMCut 30 & \bestPerformerLMCut 42.1(1.8) & \bestPerformer 30 & \bestPerformer 29.5(1.8) \\
 & 0.2, 1 & \bestPerformerCROC 30 & \bestPerformerCROC 600.7(3.6) & \bestPerformerLMCut 30 & \bestPerformerLMCut 863.4(9.7) & 0 &  & 0 &  & \bestPerformer 30 & \bestPerformer 65.1(11.1) \\
 & 0.2, \(\infty\) & \bestPerformerCROC 30 & \bestPerformerCROC 121.7(1.9) & \bestPerformerLMCut 30 & \bestPerformerLMCut 185.7(1.2) & 30 & 1537.7(72.8) & 0 &  & \bestPerformer 30 & \bestPerformer 41.7(4.9) \\
 & 0.4, 1 & \bestPerformerCROC 30 & \bestPerformerCROC 882.8(31.7) & \bestPerformerLMCut \bestPerformer 30 & \bestPerformerLMCut \bestPerformer 564.1(12.8) & 0 &  & 0 &  & 3 & 801.5(536.2) \\
 & 0.4, \(\infty\) & \bestPerformerCROC \bestPerformer 30 & \bestPerformerCROC \bestPerformer 130.9(2.5) & \bestPerformerLMCut 30 & \bestPerformerLMCut 188.1(1.3) & 0 &  & 0 &  & 3 & 665.9(308.3) \\
 & 0.6, 1 & \bestPerformerCROC 30 & \bestPerformerCROC 318.4(10.3) & \bestPerformerLMCut \bestPerformer 30 & \bestPerformerLMCut \bestPerformer 255.4(6.9) & 0 &  & 0 &  & 0 &  \\
 & 0.6, \(\infty\) & \bestPerformerCROC \bestPerformer 30 & \bestPerformerCROC \bestPerformer 25.5(0.5) & \bestPerformerLMCut 30 & \bestPerformerLMCut 39.6(0.3) & 26 & 1508.9(68.9) & 0 &  & 0 &  \\
 & 0.8, 1 & \bestPerformerCROC 30 & \bestPerformerCROC 345.5(7.4) & \bestPerformerLMCut \bestPerformer 30 & \bestPerformerLMCut \bestPerformer 267.1(8.0) & 0 &  & 0 &  & 0 &  \\
 & 0.8, \(\infty\) & \bestPerformerCROC \bestPerformer 30 & \bestPerformerCROC \bestPerformer 25.7(0.6) & \bestPerformerLMCut 30 & \bestPerformerLMCut 39.8(0.3) & 29 & 1375.1(89.9) & 0 &  & 0 &  \\
 & 1.0, 1 & \bestPerformerCROC \bestPerformer 30 & \bestPerformerCROC \bestPerformer 208.1(6.8) & \bestPerformerLMCut 30 & \bestPerformerLMCut 226.0(5.4) & 0 &  & 0 &  & 0 &  \\
 & 1.0, \(\infty\) & \bestPerformerCROC \bestPerformer 30 & \bestPerformerCROC \bestPerformer 25.6(0.6) & \bestPerformerLMCut 30 & \bestPerformerLMCut 39.6(0.3) & 30 & 1298.1(55.3) & 0 &  & 0 &  \\
\hline \multirow[c]{9}{*}{\rotatebox{90}{{\centering \tireworld (n,d,c)}}} & 4, 4, 2 & \bestPerformerCROC \bestPerformer 30 & \bestPerformerCROC \bestPerformer 1.6(0.1) & \bestPerformerLMCut \bestPerformer 30 & \bestPerformerLMCut \bestPerformer 1.8(0.1) & 30 & 4.5(0.1) & 30 & 24.2(0.7) & 30 & 24.3(0.8) \\
 & 4, 4, 4 & \bestPerformerCROC \bestPerformer 30 & \bestPerformerCROC \bestPerformer 2.4(0.0) & \bestPerformerLMCut 30 & \bestPerformerLMCut 3.2(0.0) & 30 & 8.2(0.5) & 30 & 43.0(1.0) & 30 & 36.1(2.0) \\
 & 4, 4, 6 & \bestPerformerCROC \bestPerformer 30 & \bestPerformerCROC \bestPerformer 3.1(0.0) & \bestPerformerLMCut 30 & \bestPerformerLMCut 4.6(0.1) & 30 & 10.5(0.4) & 30 & 56.0(2.4) & 30 & 42.5(1.5) \\
 & 4, 5, 2 & \bestPerformerCROC \bestPerformer 30 & \bestPerformerCROC \bestPerformer 3.9(0.2) & \bestPerformerLMCut \bestPerformer 30 & \bestPerformerLMCut \bestPerformer 4.2(0.3) & 30 & 18.7(0.8) & 30 & 119.1(5.5) & 30 & 115.3(6.3) \\
 & 4, 5, 4 & \bestPerformerCROC \bestPerformer 30 & \bestPerformerCROC \bestPerformer 5.9(0.1) & \bestPerformerLMCut 30 & \bestPerformerLMCut 7.5(0.1) & 30 & 32.4(1.0) & 30 & 239.5(9.0) & 30 & 157.0(5.3) \\
 & 4, 5, 6 & \bestPerformerCROC \bestPerformer 30 & \bestPerformerCROC \bestPerformer 7.3(0.3) & \bestPerformerLMCut 30 & \bestPerformerLMCut 10.3(0.4) & 30 & 41.7(1.0) & 30 & 310.6(14.7) & 30 & 194.5(7.1) \\
 & 4, 6, 2 & \bestPerformerCROC \bestPerformer 30 & \bestPerformerCROC \bestPerformer 9.3(0.5) & \bestPerformerLMCut \bestPerformer 30 & \bestPerformerLMCut \bestPerformer 10.1(0.6) & 30 & 90.9(4.7) & 30 & 840.2(47.1) & 30 & 937.7(45.7) \\
 & 4, 6, 4 & \bestPerformerCROC \bestPerformer 30 & \bestPerformerCROC \bestPerformer 13.9(0.3) & \bestPerformerLMCut 30 & \bestPerformerLMCut 17.8(0.3) & 30 & 187.0(5.9) & 13 & 1656.6(76.8) & 30 & 1281.6(64.0) \\
 & 4, 6, 6 & \bestPerformerCROC \bestPerformer 30 & \bestPerformerCROC \bestPerformer 18.2(0.4) & \bestPerformerLMCut 30 & \bestPerformerLMCut 24.9(0.7) & 30 & 225.1(10.0) & 4 & 1424.3(88.9) & 30 & 1238.9(81.6) \\
\hline \end{tabular}}
\caption{For the benchmark problems, we show each planner and heuristic's coverage (out of 30) and over the converged runs the mean runtime (secs) with 95\% C.I.
For each problem, the fastest planner and heuristic pairs are in bold.}
\label{tab:main-results}
\end{table*}

The results of our experiments are presented in \cref{tab:main-results}.
We give each algorithm's coverage (the number of instances solved within its time and memory limits), and over the instances where the algorithm converged we give its mean runtime (and 95\% C.I.).
We highlight the fastest algorithms, where an algorithm is considered the fastest for a particular problem if it has the maximum coverage and then its mean time plus 95\% C.I. is less than the next best algorithm's mean time minus 95\% C.I.
We now identify the best performing algorithm per domain.

\paragraph{\sar, \exbw, and \tireworld.}
\carl dominates in these domains: either \(\carl(\lambdaroc)\) or \(\carl(\lambdalmcut)\) is the best performing planner and, with exception of \exbw (1,5,.1) and (3,6,.91), the second best performing planning is also \carl.
Moreover, \carl can obtain substantially larger coverage with the extreme case of \exbw (6,8,.3) in which \carl using both heuristics obtained full coverage while all other planners had zero coverage.
Comparing only entries with full coverage in these domains, \carl has \(10 \times\), \(19 \times\), and \(42 \times\) speedup on average w.r.t. \idual(\croc), \idual(\iplmcut) and \iidual, respectively. %
Between \(\carl(\lambdaroc)\) and \(\carl(\lambdalmcut)\), there is no clear winner on \sar; \(\carl(\lambdalmcut)\) dominates on \exbw; and \(\carl(\lambdaroc)\) dominates on \tireworld.

\paragraph{\elevators.}
With one elevator \idual dominates, and with two elevators \carl dominates.
With one elevator, \carl fails to solve some instances for one of two reasons: (1) \carl's coordinate search fails and falls back on the subgradient method, which is significantly slower than coordinate search; (2) \lambdassps are too difficult to solve in some cases, which suggests the heuristic does not give adequate guidance.
Regarding (1), we note that these \elevators instances are the only cases in our benchmarks where coordinate search fails and \carl is forced to fall back on the subgradient method to ensure completeness (see \cref{sec:carl-step-1}).
Here, the subgradient method is significantly slower than coordinate search and failed to converge in time.
However, we reiterate that falling back to the subgradient method ensures that \carl is complete, and given enough time and memory \carl would solve the instances.
With two elevators, the problem structure changes so that coordinate search succeeds and (1) does not occur, and any instances where \carl fails to find a solution is because of (2), that is, the coordinate search's subproblems take too long to solve.
This contrasts \idual and \iidual, which are not affected by problem structure but scale worse with the larger search spaces as the second elevator is introduced.

\paragraph{\parc.}
\iidual dominates when the constraint over failure is tighter (\(f \leq 0.2\)) and \carl dominates when the constraint is looser (\(f \geq 0.4\)).
\carl struggles with the tight constraints because its heuristic fails to take secondary-cost constraints into account.
This is supported by the result that \carl becomes competitive with \iidual on the tightly constrained problems when used with \croc, a heuristic that takes secondary-cost constraints into account (see \cref{sec:more-results}).\footnote{\carl(\croc) is not in the main results because \croc is inadmissible for \lambdassps, so the algorithm has no optimality guarantee.}
\Iidual has the opposite weakness: its heuristic is very informative for highly constrained problems and it prunes a lot of the search space but, as the constraints become looser, its heuristic becomes less effective.
On problems with the loosest constraints, namely \((0.6,1)\), \((0.8,1)\), and \((1.0, 1)\), \carl completely dominates by achieving full coverage while the baselines obtain zero coverage.

\paragraph{What makes \carl so effective on our benchmarks?}
The part of \carl that takes the most time is solving \lambdassps and, over the solved instances, \carl only solves a mean and maximum of \(6.7\) and \(30\) \lambdassps in its search.
We identify two domain features that make coordinate search efficient and let \carl solve so few subproblems: single secondary-cost constraint and sparse $\vlambda$.
In the former, coordinate search only has to solve a single coordinate and it is guaranteed to converge to \(\vlambda^*\).
\sar, \exbw, and \parc problems with \(u = \infty\) have a single constraint.
It is convenient if \(\vlambda^* = 0\) because \(\vlambda=0\) is the initial guess and coordinate search then only has to verify that its candidate is optimal.
\tireworld has \(\vlambda^* = 0\) for almost all its problems.
More generally, coordinate search tends to work well when \(\vlambda^*\) is sparse because it is inexpensive to determine these, but this trend is not guaranteed and depends on the search space structure.
Over the solved instances, \(\vlambda^*\) is quite sparse having a mean and maximum of \(0.62\) and \(2\) non-zero entries.
Note that if \(\C[i](\p^*) < \ubC{i}\), i.e., a secondary-cost constraint is satisfied with some slack, then \(\lambda^*_i = 0\) by complementary slackness, which suggests that many problems will have a sparse \(\vlambda^*\).
In contrast, we have seen that particular problems are structured such that coordinate search fails to find \(\vlambda^*\), e.g., some \elevators instances with a single elevator.
Fortunately, these seem to appear infrequently.
We emphasise that these benchmarks were taken from the existing literature on CSSPs, and were not hand-picked to accommodate \carl.
This may suggest that problems are more often amenable to \carl than not.

\section{Conclusion and Future Work}

We introduced \carl, a novel algorithm for finding optimal, potentially stochastic, policies for CSSPs.
It works by solving a sequence of scalarised SSPs, called \lambdassps, with efficient heuristic
search algorithms for SSPs, and combining the optimal policies of the final \lambdassp solved into an optimal policy for the original CSSP.
\carl is the first heuristic search algorithm over the primal space for finding optimal CSSP policies and, although \carl follows the Lagrangian ascent framework from OR, this paper presents novel methods for solving \lambdassps efficiently, adapting heuristics to \lambdassps and extracting stochastic policies.
\carl outperforms the state-of-the-art algorithms on nearly all our benchmark problems, solving \(50\%\) more instances and, for some problems, it solves all instances while the state-of-the-art solve none.
Over the problems where all algorithms solved all problems, \carl offers an average \(10\times\) speedup.

In terms of future work, we believe more investigation is required into optimisation procedures for finding \(\vlambda^*\).
\carl uses coordinate search, which works well and is significantly faster than the plain subgradient method, but it is unsatisfying that it can not guarantee an optimal solution and it may not scale well with many secondary-cost constraints.
We also believe it is important to better understand the problems themselves and what features make them more or less amenable to \carl's search procedure.

\section{Acknowledgements}

This research project was undertaken with the assistance of resources and services from the National Computational Infrastructure (NCI), which is supported by the Australian Government.
Johannes Schmalz received funding from DFG grant 389792660 as part of TRR 248 (see \url{https://www.perspicuous-computing.science}).

\clearpage

\appendix

\section{CSSP Example}\label{sec:cssp-examples}

Here, we present an example of a CSSP.

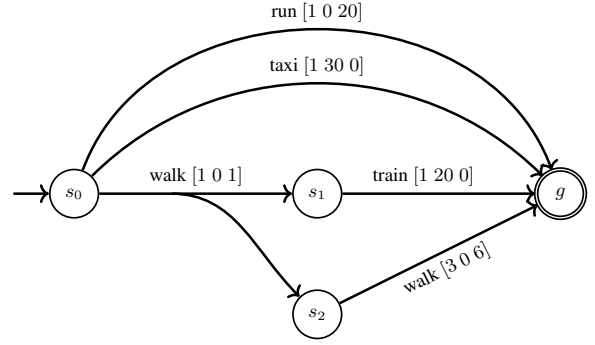
\begin{figure}[t!]
\center
\scalebox{0.8}{

\newcommand{\scaleX}{2} %
\newcommand{\scaleY}{1} %
\begin{tikzpicture}
\begin{scope}[every node/.style={circle, thick, draw, minimum size=0.8cm, inner sep=0.04cm}]
  \coordinate (-1) at (-1, 0);
	\node (0) at (0*\scaleX, 0*\scaleY) {\(s_0\)};
	\node (1) at (2*\scaleX, 0*\scaleY) {\(s_1\)};
	\node (2) at (2*\scaleX, -2*\scaleY) {\(s_2\)};
	\node[double] (g) at (4*\scaleX, 0*\scaleY) {\(g\)};
\end{scope}
\begin{scope}
  \coordinate (fork-walk) at ($(0)!0.4!(1)$);
\end{scope}
\begin{scope}[every edge/.style={draw=black, very thick}]
\path [->] (-1) edge (0);
\path [-] (0) edge[] node[above] {\hspace{2cm}walk \([1 \; 0 \; 1]\)} (fork-walk);
\path [->] (fork-walk) edge[] (1);
\path [->] (fork-walk) edge[out=0] (2);
\path [->] (1) edge[] node[above] {\hspace{-0.5cm}train \([1 \; 20 \; 0]\)} (g);
\path [->] (2) edge[] node[below, rotate=26.4] {walk \([3 \; 0 \; 6]\)} (g);
\path [->] (0) edge[out=70, in=180-70, looseness=1.1] node[above] {run \([1 \; 0 \; 20]\)} node[below] {} (g);
\path [->] (0) edge[out=45, in=180-45] node[above] {taxi \([1 \; 30 \; 0]\)} (g);
\end{scope}
\end{tikzpicture}
\undef{\scaleX} %
\undef{\scaleY} %

} %
\caption{The CSSP associated with the ``Getting to Work'' example.}
\label{fig:cssp-getting-to-work-example}
\end{figure}

\paragraph{Getting to Work.}
The agent needs to make its way from home to work.
It can run, use a taxi, or walk to the train station where it can try to take the train.
The train is cancelled with \(50\%\) probability.
Each action has a cost vector \([t \; p \; e]\) in terms of time (\(t\)), price (\(p\)), and personal effort (\(e\)).
The agent's task is to get to work in minimal time \st price \(\leq 15\) and effort \(\leq 10\) over expectation.
The CSSP is shown in \cref{fig:cssp-getting-to-work-example}.
This problem has three proper deterministic policies:
\begin{itemize}
\item \(\p_{\text{run}} = \{ \sZ \mapsto \text{run} \}\),
\item \(\p_{\text{taxi}} = \{ \sZ \mapsto \text{taxi} \}\),
\item \(\p_{\text{train}} = \{ \sZ \mapsto \text{walk}, \s_1 \mapsto \text{train}, \s_2 \mapsto \text{train} \}\).
\end{itemize}
This problem has a unique optimal policy \(\p^*\) which is stochastic with \(\p^*(\sZ, \text{run}) = 0.5\) and \(\p^*(\sZ, \text{taxi}) = 0.5\) where
\begin{itemize}
\item \(\C[0](\p^*) = 1\) (time),
\item \(\C[1](\p^*) = 15 \leq 15\) (price),
\item \(\C[2](\p^*) = 10 \leq 10\) (effort).
\end{itemize}
Note that \(\p^*\) can be expressed as \(\p^* = 0.5 \p_{\text{run}} + 0.5 \p_{\text{taxi}}\).

\section{Visualisation of \(\lagrange(\vlambda)\)}\label{sec:visualisation-of-L}

In this section, we visualise the surface of \(\lagrange(\vlambda)\) for two different CSSPs.

\begin{figure}[t!]
\includegraphics[width=\linewidth]{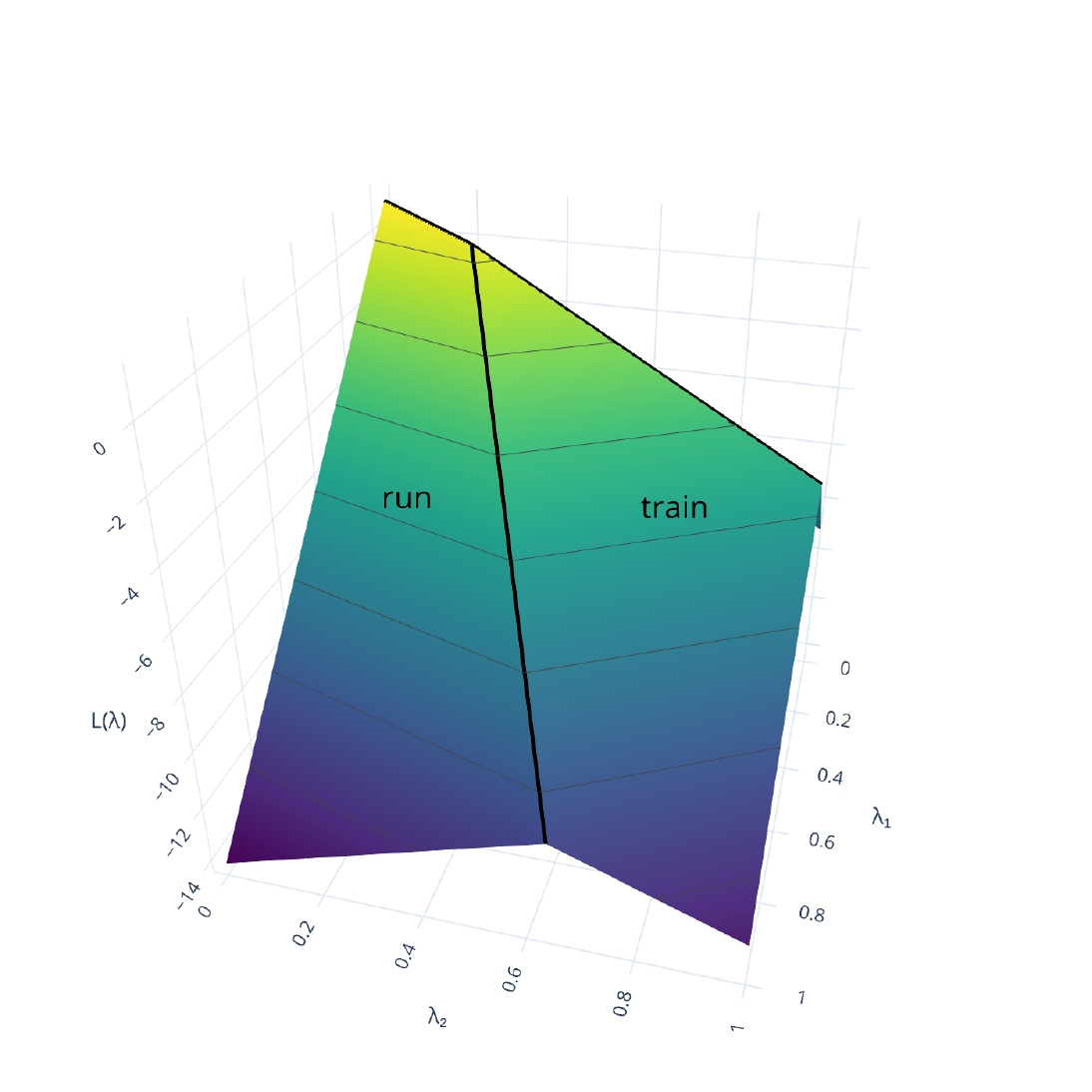}
\caption{\(\lagrange(\vlambda)\) for the ``Getting To Work'' example.}
\label{fig:L-plot-getting-to-work}
\end{figure}

\paragraph{Getting to Work.}
First, we consider the ``Getting to Work'' example from \cref{sec:cssp-examples}.
\Cref{fig:L-plot-getting-to-work} shows this problem's \(\lagrange(\vlambda)\) as \vlambda varies.
Each linear surface of the plot corresponds to one of the deterministic policies; we have labelled the linear surfaces associated with \(\p_{\text{run}}\) and \(\p_{\text{train}}\) with ``run'' and ``train'' respectively, and \(\p_{\text{taxi}}\) is obscured.
A linear surface indicates that the corresponding policy is optimal for the \lambdassp at the relevant values of \vlambda; edges and vertices indicate that multiple deterministic policies are optimal at the \lambdassp.
In particular, note that at \(\vlambda^* = 0\) we have an edge between \(\p_{\text{run}}\) and \(\p_{\text{taxi}}\), and \(\p_{\text{train}}\) does not intersect there; this indicates that the optimal (in this case stochastic) policy consists of \(\p_{\text{run}}\) and \(\p_{\text{taxi}}\), which we know to be true.
This example is not interesting in terms of coordinate search, because \(\vlambda^* = 0\) is found in the first step.

\paragraph{Interesting CSSP for Coordinate Search.}

\begin{figure}[t!]
\centering
\scalebox{0.8}{

\newcommand{\scaleX}{2} %
\newcommand{\scaleY}{1} %
\begin{tikzpicture}
\begin{scope}[every node/.style={circle, thick, draw, minimum size=0.8cm, inner sep=0.04cm}]
        \coordinate (-1) at (-1, 0);
        \node (0) at (0, 0) {\(\sZ\)};
        \node (1) at (\scaleX*2, 0) {\(\s_1\)};
        \node[double] (g) at (\scaleX*4, 0) {\(\s_g\)};
\end{scope}
\begin{scope}[every edge/.style={draw=black, very thick}]
  \path[->] (-1) edge (0);
	\path[->] (0) edge[out=90,in=180-90] node[above] {\([1 \; 40 \; 40]\)} node[below] {\(\ac_0\)} (g);
	\path[->] (0) edge[out=40,in=180-40] node[above] {\([5 \; 5 \; 5]\)} node[below] {\(\ac_1\)} (1);
	\path[->] (0) edge[] node[above] {\([3 \; 10 \; 0]\)} node[below] {\(\ac_2\)} (1);
	\path[->] (0) edge[out=-40,in=180+40] node[above] {\([1 \; 0 \; 20]\)} node[below] {\(\ac_3\)} (1);
	\path[->] (1) edge[out=40,in=180-40] node[above] {\([1 \; 20 \; 0]\)} node[below] {\(\ac_4\)} (g);
	\path[->] (1) edge[out=-40,in=180+40] node[above] {\([1 \; 0 \; 20]\)} node[below] {\(\ac_5\)} (g);
\end{scope}
\end{tikzpicture}
\undef{\scaleX} %
\undef{\scaleY} %
}

\caption{A CSSP that is interesting for coordinate search.
It has \(\ubC{1} = \ubC{2} = 15\).}
\label{fig:cssp-interesting-coordinate-search}
\end{figure}
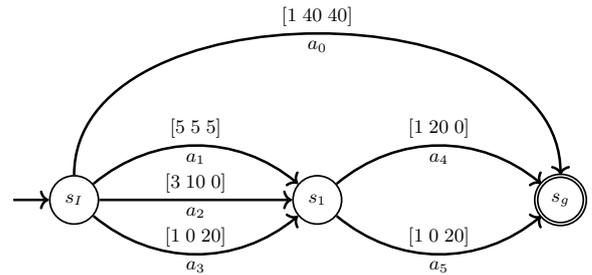

\begin{figure}[t!]
\includegraphics[width=\linewidth]{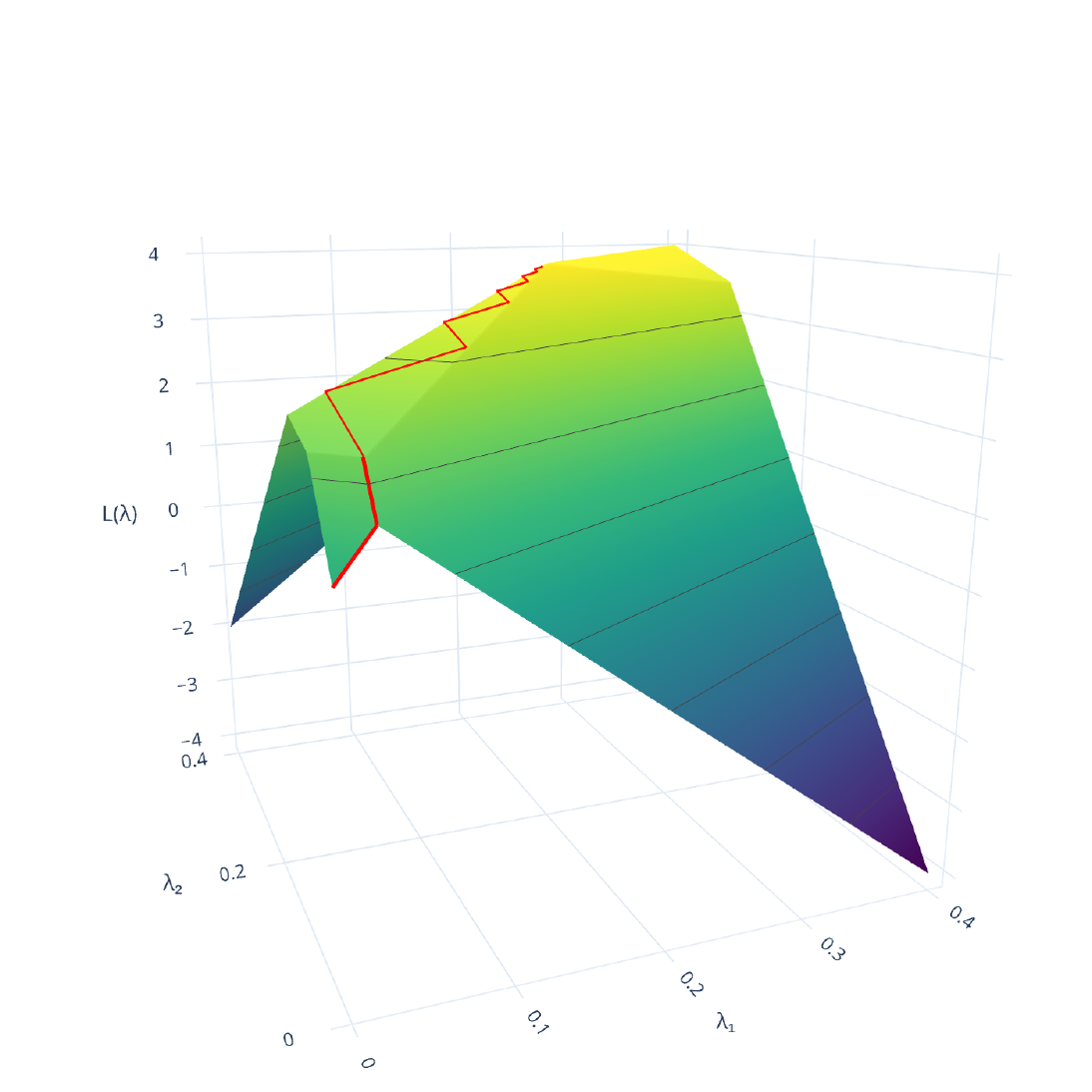}
\caption{\(\lagrange(\vlambda)\) for the ``Interesting CSSP for Coordinate Search'' example.}
\label{fig:L-plot-interesting-coordinate-search}
\end{figure}

Consider the CSSP in \cref{fig:cssp-interesting-coordinate-search}, with \(\ubC{1} = \ubC{2} = 15\).
Its optimal policy \(\p^*\) is stochastic with
\begin{itemize}
\item \(\p^*(\sZ, \ac_2) = 1\)
\item \(\p^*(\s_1, \ac_4) = \frac{1}{4} \quad \p^*(\s_1, \ac_5) = \frac{3}{4}\)
\end{itemize}
and it has \(\C[0](\p^*) = 4\) and \(\C[1](\p^*) = \C[2](\p^*) = 15\).
\Cref{fig:L-plot-interesting-coordinate-search} shows this CSSP's \(\lagrange(\vlambda)\) as \vlambda varies.
As before, each linear surface corresponds to a deterministic policy that is optimal for the relevant \lambdassp.
We visualise the values of \vlambda considered by coordinate search with the red lines across the surface, i.e., the line starts at the initial value of \(\vlambda = 0\), then moves to \(\vlambda = [0.025 \; 0]\), then \(\vlambda = [0.025 \; 0.025]\), and so on, eventually converging to \(\vlambda^* = [0.2 \; 0.2]\).
Recall that coordinate search only changes a single coordinate per step, which gives the red lines' ``stair-case'' shape.

\section{Pathological Example for Coordinate Search}\label{sec:coordinate-search-edge-case}

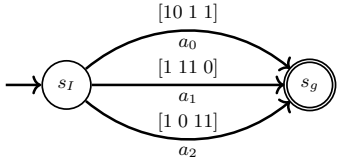
\begin{figure}[t!]

\center
\scalebox{0.8}{

\newcommand{\scaleX}{1} %
\newcommand{\scaleY}{1} %
\begin{tikzpicture}
\begin{scope}[every node/.style={circle, thick, draw, minimum size=0.8cm, inner sep=0.04cm}]
        \coordinate (-1) at (-1, 0);
        \node (0) at (0, 0) {\(\sZ\)};
        \node[double] (g) at (\scaleX*4, 0) {\(\s_g\)};
\end{scope}
\begin{scope}[every edge/.style={draw=black, very thick}]
  \path[->] (-1) edge (0);
	\path[->] (0) edge[out=40,in=180-40] node[above] {\([10 \; 1 \; 1]\)} node[below] {\(\ac_0\)} (g);
	\path[->] (0) edge[] node[above] {\([1 \; 11 \; 0]\)} node[below] {\(\ac_1\)} (g);
	\path[->] (0) edge[out=-40,in=180+40] node[above] {\([1 \; 0 \; 11]\)} node[below] {\(\ac_2\)} (g);
\end{scope}
\end{tikzpicture}
\undef{\scaleX} %
\undef{\scaleY} %
}
\caption{A pathological CSSP where coordinate search fails to find \(\vlambda^*\). It has \(\ubC{1} = \ubC{2} = 1\).}
\label{fig:cssp-pathological-example-for-coordinate-search}
\end{figure}

\begin{figure}[t!]
\includegraphics[width=\linewidth]{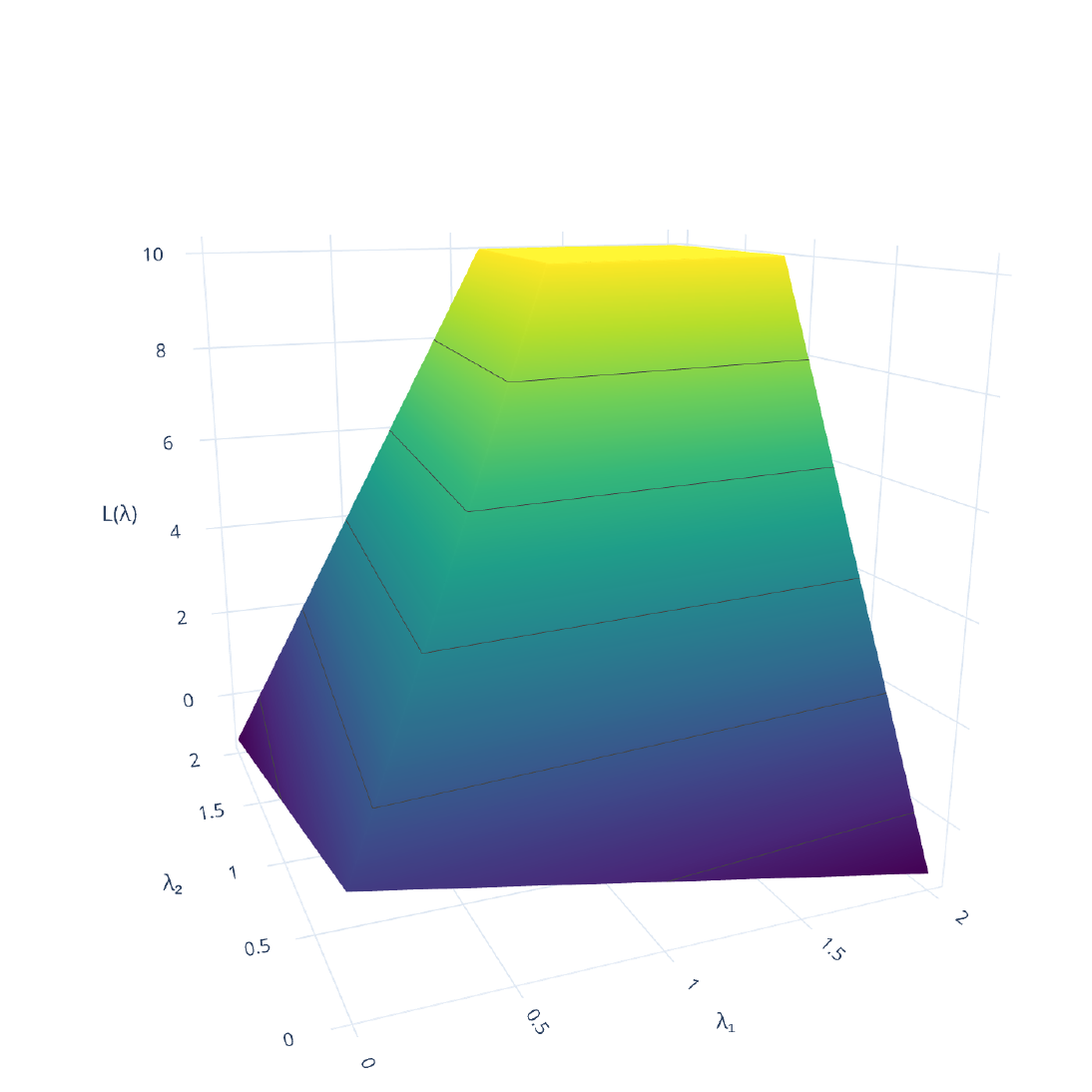}
\caption{Surface of \(\lagrange(\vlambda)\) for the pathological CSSP where coordinate search fails to find \(\vlambda^*\).}
\label{fig:L-plot-pathological-coordinate-search}
\end{figure}

There are non-smooth problems where coordinate search fails.
For example, consider the CSSP in \cref{fig:cssp-pathological-example-for-coordinate-search} with two secondary-cost constraints \(\ubC{1} = \ubC{2} = 1\), i.e., a policy \p is feasible if \(\C[1](\p) \leq 1\) and \(\C[2](\p) \leq 1\).
The CSSP has three actions that lead deterministically from \sZ to the goal with costs
\begin{itemize}
\item \(\vecC(\ac_0) = [10 \; 1 \; 1]\),
\item \(\vecC(\ac_1) = [1 \; 11 \; 0]\), and
\item \(\vecC(\ac_2) = [1 \; 0 \; 11]\),
\end{itemize}
inducing the three deterministic policies \(\p_0, \p_1, \p_2\) with \(\p_i(\sZ) = \ac_i\) for \(i \in \{0, 1, 2\}\).
Taking into account the terminal costs, the policy costs are
\begin{itemize}
\item \(\C_{\vlambda}(\p_0) = 10 + \lambda_1 (1 - \ubC{1}) + \lambda_2 (1 - \ubC{2}) = 10\),
\item \(\C_{\vlambda}(\p_1) = 1 + \lambda_1(11 - \ubC{1}) + \lambda_2(0 - \ubC{2}) = 1 + 10 \lambda_1 - \lambda_2\), and
\item \(\C_{\vlambda}(\p_2) = 1 + \lambda_1(0 - \ubC{1}) + \lambda_2(11 - \ubC{2}) = 1 - \lambda_1 + 10 \lambda_2\).
\end{itemize}
\(\lagrange(\vlambda)\) selects the cheapest one of these, i.e.,
\(
\lagrange(\vlambda) = \min \{
10,
1 + 10\lambda_1 - \lambda_2,
1 - \lambda_1 + 10\lambda_2
\}.
\)
This surface is shown in \cref{fig:L-plot-pathological-coordinate-search}; note there are three linear surfaces corresponding to the three deterministic policies (the flat one is \(\p_0\), and the sloped ones are \(\p_1\) and \(\p_2\)).
If we consider \(\vlambda = 0\) as a starting point then \(\lagrange([0 \; 0]) = 1\) and we can see that there is no way to get a larger \(\lagrange(\vlambda)\) by moving in a single coordinate.
Algebraically,
\[
\lagrange([x \; 0]) = \min\{ 10, 1+10x, 1-x \} = 1 - x < 1
\]
and symmetrically
\[
\lagrange([0 \; x]) = \min\{ 10, 1-x, 1+10x \} = 1 - x < 1\]
for all \(x \in \Reals_{>0}\).
Thus, coordinate search gets stuck at \(\vlambda = [0 \; 0]\) with \(\lagrange(\vlambda) = 1\) and fails to progress to an optimal solution such as \(\vlambda^* = [2 \; 2]\) with \(\lagrange(\vlambda^*) = 10\).

\section{Bellman Backups over Vector Value Function}\label{sec:vector-bellman-backups}

Instead of using a scalar value function \(\V : \Ss \to \Reals_{\geq 0}\), we use a vector value function \(\vecV : \Ss \to \Reals^{n+1}_{\geq 0}\), with entries for each cost function.
This modification still solves the \lambdassp SSP by projecting the vector value function onto \(\V_{\vlambda}(\s) = [1 \; \vlambda] \cdot \vecV(\s)\), but we now apply Bellman backups on \vecV directly.
We write \(\vecQ(\s,\ac) = \vecC(\ac) + \sum_{\s' \in \Ss} \pr(\s'|\s,\ac) \vecV(\s')\), then the \(\lambda\) modified Bellman backups are given as follows:
\begin{definition}[\(\lambda\) Bellman backup]\label{def:lambda-bellman-backup}
Given the vector value function \(\vecV : \Ss \to \Reals_{\geq 0}^{n+1}\) and a scalarisation \(\vlambda \in \Reals_{\geq 0}^{n}\), a vector Bellman backup applied to state \s sets
\[ \vecV(\s) \gets \vecQ(\s,\ac_{\min}) \quad\text{ for }\quad \ac_{\min} \in \argmin_{\ac \in \A(\s)} \Big( [1 \; \vlambda] \cdot \vecQ(\s,\ac) \Big).\]
\end{definition}
By applying \(\lambda\) Bellman backups, we can solve \lambdassp and evaluate its policy for each cost function \(\C[i](\p^*_{\vlambda})\) at the same time.
This idea was used by \citeintext{Hong2023:DetPiForCSSPs}.

The \(\lambda\) Bellman backups introduce two technical issues.
First, to ensure that each \(\C[i](\p^*_{\vlambda})\) is evaluated accurately, we must redefine the Bellman residual so that each value function converges, otherwise the backups may converge to an optimal vector value function for the \lambdassp without the scalar value functions converging.
We can do this by considering \(\max_{i=0}^{n} |\V_i(\s) - \Q(\s,\ac_{\min})|\) for \s's \(\ac_{\min}\).
The second technical issue follows from the redefined Bellman residual, because it is possible to get stuck in cycles where actions with similar scalarised \qvalues change individual value functions significantly, so that the new residual never goes to 0.
To address this, we need to use tie-breaking rules, e.g., if two actions have costs within \(\epsilon\) of each other, iterate through the value function indices and pick the action that has a smaller \qvalue first.

\section{\Strongeconsistency vs \econsistency}\label{sec:strong-econsistency-vs-econsistency}

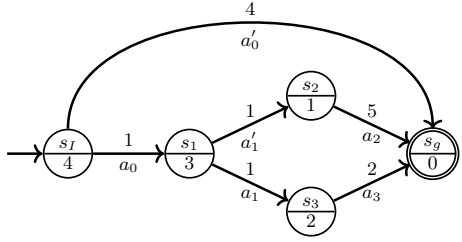
\begin{figure}[t!]
\centering
\scalebox{0.8}{
\newcommand{\scaleX}{1} %
\newcommand{\scaleY}{1.2} %
\begin{tikzpicture}
\begin{scope}[every node/.style={circle split, thick, draw, minimum size=0.8cm, inner sep=0.04cm}]
        \coordinate (-1) at (-1, 0);
        \node (0) at (\scaleX*0, \scaleY*0) {\(\sZ\) \nodepart{lower} \(4\)};
        \node (1) at (\scaleX*2, \scaleY*0) {\(\s_1\) \nodepart{lower} \(3\)};
        \node (2) at (\scaleX*4, \scaleY*0.8) {\(\s_2\) \nodepart{lower} \(1\)};
        \node (3) at (\scaleX*4, \scaleY*-0.8) {\(\s_3\) \nodepart{lower} \(2\)};
        \node[double] (g) at (\scaleX*6, \scaleY*0) {\(\s_g\) \nodepart{lower} \(0\)};
\end{scope}
\begin{scope}[every edge/.style={draw=black, very thick}]
        \path [->] (-1) edge (0);
        \path [->] (0) edge[out=90, in=180-90, looseness=1] node[above] {\(4\)} node[below] {\(\ac'_0\)} (g);
        \path [->] (0) edge[] node[above] {\(1\)} node[below] {\(\ac_0\)} (1);
        \path [->] (1) edge[] node[above] {\(1\)} node[below] {\(\ac'_1\)} (2);
        \path [->] (1) edge[] node[above] {\(1\)} node[below] {\(\ac_1\)} (3);
        \path [->] (2) edge[] node[above] {\(5\)} node[below] {\(\ac_2\)} (g);
        \path [->] (3) edge[] node[above] {\(2\)} node[below] {\(\ac_3\)} (g);
\end{scope}
\end{tikzpicture}}
\undef{\scaleX} %
\undef{\scaleY} %

\caption{An SSP with a value function that is \econsistent but not \strongeconsistent, and fails to describe all optimal policies.}
\label{fig:strong-econsistency-example}
\end{figure}

For unconstrained SSPs, if a value function \V is \econsistent~\cite{Bonet2003:hdp}, then it is guaranteed to induce a deterministic optimal policy (as \(\epsilon \to 0)\); however, it may not induce all optimal deterministic policies.
To produce all optimal deterministic policies, we require \strongeconsistency.
We give an example where \econsistency fails to capture all optimal deterministic policies: consider the SSP in \cref{fig:strong-econsistency-example}, with the value function \V specified in the bottom of each node.
The SSP has two optimal policies:
\begin{itemize}
\item \(\p^*_0 = \{ \sZ \mapsto \ac'_0 \}\)
\item \(\p^*_1 = \{ \sZ \mapsto \ac_0, \s_1 \mapsto \ac_1, \s_3 \mapsto \ac_3 \}\).
\end{itemize}
\V has two greedy policies: the optimal \(\p^*_0\) and a suboptimal policy
\begin{itemize}
\item \(\p' = \{ \sZ \mapsto \ac_0, \s_1 \mapsto \ac'_1, \s_2 \mapsto \ac_2 \}\).
\end{itemize}
\V is \econsistent \wrt the greedy policy \(\p^*_0\), and \V is admissible; so, \V could be produced by an optimal SSP algorithm such as \cgilao~\cite{Schmalz2024:cgilao} (which is what we use to solve our \lambdassp SSPs).
Importantly, \V is not \econsistent \wrt the other greedy policy \(\p'\) because \(\residual(\s_2) = |1 - 5| > \epsilon\) (assume small \(\epsilon\) such as \(0.0001\)), and therefore \V is not \strongeconsistent.
This has the effect that \V fails to capture the second optimal policy \(\p^*_1\), i.e., \(\p^*_1\) is not a greedy policy \wrt \V.

\section{More Results}\label{sec:more-results}

We present an extended table of results in \cref{tab:extended-results}.
It has the same entries as \cref{tab:main-results}, and additionally includes \carl with ideal-point heuristics and \croc.
Recall that \croc is not admissible for \carl's \lambdassps, so it provides no guarantee of optimality nor convergence.
Nevertheless, in our experiments, wherever \(\carl(\croc)\) has a coverage entry, it found the optimal solution.

\begin{table*}[t]
\resizebox{\textwidth}{!}{
\setlength\tabcolsep{2px}

\begin{tabular}{|ll| S[table-format=2] S[table-format=4.1(3.1)]| S[table-format=2] S[table-format=4.1(3.1)]| S[table-format=2] S[table-format=4.1(3.1)]| S[table-format=2] S[table-format=4.1(3.1)]| S[table-format=2] S[table-format=4.1(3.1)]| S[table-format=2] S[table-format=4.1(3.1)]| S[table-format=2] S[table-format=4.1(3.1)]| S[table-format=2] S[table-format=4.1(3.1)]|llllll}
\hline
 & & \multicolumn{10}{c|}{\carl} & \multicolumn{4}{c|}{\idual} & \multicolumn{2}{c|}{\iidual} \\
 & & \multicolumn{2}{c}{\lambdaroc} & \multicolumn{2}{c}{\iproc} & \multicolumn{2}{c}{\croc} & \multicolumn{2}{c}{\lambdalmcut} & \multicolumn{2}{c|}{\iplmcut} & \multicolumn{2}{c}{\croc} & \multicolumn{2}{c|}{\iplmcut} & \multicolumn{2}{c|}{} \\
\hline
\multirow[c]{6}{*}{\rotatebox{90}{\parbox{2cm}{\centering \sar \\ (n, d, r)}}} & (4, 3, 0.75) & \bestPerformerCROC \bestPerformer 30 & \bestPerformerCROC \bestPerformer 5.6(1.4) & 30 & 11.5(2.3) & 30 & 12.1(2.4) & \bestPerformerLMCut \bestPerformer 30 & \bestPerformerLMCut \bestPerformer 4.3(1.0) & 30 & 9.2(1.8) & 30 & 104.1(51.0) & 30 & 45.8(21.1) & 30 & 69.2(56.4) \\
 & (4, 4, 0.75) & \bestPerformerCROC \bestPerformer 30 & \bestPerformerCROC \bestPerformer 41.9(5.3) & 30 & 75.5(7.6) & 30 & 79.5(8.3) & \bestPerformerLMCut \bestPerformer 30 & \bestPerformerLMCut \bestPerformer 34.1(4.2) & 30 & 65.3(7.6) & 5 & 1345.7(534.1) & 9 & 979.8(296.6) & 8 & 495.5(417.8) \\
 & (5, 3, 0.50) & \bestPerformerCROC \bestPerformer 30 & \bestPerformerCROC \bestPerformer 4.9(0.9) & 30 & 9.0(1.6) & 30 & 9.3(1.7) & \bestPerformerLMCut \bestPerformer 30 & \bestPerformerLMCut \bestPerformer 4.8(1.1) & 30 & 10.0(2.4) & 30 & 25.1(8.5) & 30 & 13.1(4.6) & 30 & 11.1(4.0) \\
 & (5, 3, 0.75) & \bestPerformerCROC \bestPerformer 30 & \bestPerformerCROC \bestPerformer 12.3(2.5) & 30 & 26.2(5.8) & 30 & 27.9(6.0) & \bestPerformerLMCut \bestPerformer 30 & \bestPerformerLMCut \bestPerformer 13.3(2.9) & 30 & 33.1(9.0) & 30 & 253.8(110.6) & 30 & 113.3(46.0) & 30 & 168.0(105.9) \\
 & (5, 4, 0.50) & \bestPerformerCROC \bestPerformer 30 & \bestPerformerCROC \bestPerformer 30.8(6.8) & \bestPerformer 30 & \bestPerformer 52.3(9.5) & \bestPerformer 30 & \bestPerformer 54.7(9.8) & \bestPerformerLMCut \bestPerformer 30 & \bestPerformerLMCut \bestPerformer 36.2(7.9) & \bestPerformer 30 & \bestPerformer 70.5(12.9) & 17 & 695.0(268.5) & 24 & 550.6(166.1) & 27 & 637.8(213.1) \\
 & (5, 4, 0.75) & \bestPerformerCROC \bestPerformer 30 & \bestPerformerCROC \bestPerformer 126.6(45.2) & \bestPerformer 30 & \bestPerformer 270.0(80.7) & \bestPerformer 30 & \bestPerformer 282.9(83.9) & \bestPerformerLMCut \bestPerformer 30 & \bestPerformerLMCut \bestPerformer 142.5(49.3) & \bestPerformer 30 & \bestPerformer 353.9(105.7) & 6 & 930.9(369.3) & 6 & 499.2(226.4) & 8 & 349.1(288.1) \\
\hline \multirow[c]{7}{*}{\rotatebox{90}{\parbox{2cm}{\centering \elevators \\ (e,w,h)}}} & (1, 1, 1) & \bestPerformerCROC 30 & \bestPerformerCROC 1.6(0.6) & \bestPerformerCROC 30 & \bestPerformerCROC 2.0(0.4) & \bestPerformerCROC 30 & \bestPerformerCROC 2.1(0.3) & 30 & 1.2(0.3) & 30 & 1.4(0.3) & \bestPerformerCROC 30 & \bestPerformerCROC 1.5(0.1) & \bestPerformerLMCut \bestPerformer 30 & \bestPerformerLMCut \bestPerformer 0.8(0.1) & 30 & 6.2(0.6) \\
 & (1, 1, 2) & 29 & 50.9(26.8) & 29 & 42.7(12.6) & 29 & 42.2(11.9) & 29 & 35.0(12.7) & 29 & 44.4(14.1) & \bestPerformerCROC 30 & \bestPerformerCROC 25.3(1.0) & \bestPerformerLMCut \bestPerformer 30 & \bestPerformerLMCut \bestPerformer 21.1(2.6) & 30 & 280.0(46.5) \\
 & (1, 2, 1) & 25 & 14.2(5.6) & 25 & 17.7(3.7) & 25 & 17.7(3.7) & 25 & 13.1(4.2) & 25 & 15.8(4.3) & \bestPerformerCROC 30 & \bestPerformerCROC 13.9(0.9) & \bestPerformerLMCut \bestPerformer 30 & \bestPerformerLMCut \bestPerformer 9.8(1.4) & 30 & 174.2(38.4) \\
 & (1, 2, 2) & 22 & 313.3(150.2) & 22 & 399.8(99.9) & 22 & 407.9(102.0) & 22 & 336.9(115.8) & 22 & 460.1(118.5) & \bestPerformerCROC \bestPerformer 30 & \bestPerformerCROC \bestPerformer 340.1(35.7) & \bestPerformerLMCut \bestPerformer 30 & \bestPerformerLMCut \bestPerformer 392.5(72.1) & 1 & 989.2 \\
 & (2, 1, 1) & \bestPerformerCROC \bestPerformer 30 & \bestPerformerCROC \bestPerformer 104.5(23.2) & \bestPerformerCROC \bestPerformer 30 & \bestPerformerCROC \bestPerformer 131.7(26.2) & \bestPerformerCROC \bestPerformer 30 & \bestPerformerCROC \bestPerformer 126.9(25.1) & \bestPerformerLMCut \bestPerformer 30 & \bestPerformerLMCut \bestPerformer 121.8(26.7) & \bestPerformerLMCut \bestPerformer 30 & \bestPerformerLMCut \bestPerformer 149.7(31.5) & 23 & 670.6(173.7) & 26 & 576.1(181.4) & 1 & 1799.7 \\
 & (2, 1, 2) & \bestPerformerCROC \bestPerformer 20 & \bestPerformerCROC \bestPerformer 1321.5(161.1) & 4 & 1281.0(114.2) & 5 & 1309.1(230.8) & \bestPerformerLMCut 3 & \bestPerformerLMCut 1344.2(341.5) & 0 &  & 1 & 1799.6 & 1 & 1799.1 & 0 &  \\
 & (2, 2, 1) & \bestPerformerCROC \bestPerformer 28 & \bestPerformerCROC \bestPerformer 569.2(131.7) & \bestPerformerCROC \bestPerformer 28 & \bestPerformerCROC \bestPerformer 830.8(160.8) & \bestPerformerCROC \bestPerformer 28 & \bestPerformerCROC \bestPerformer 796.4(158.7) & \bestPerformerLMCut 23 & \bestPerformerLMCut 890.5(164.2) & 18 & 1157.4(180.4) & 5 & 1608.9(229.1) & 5 & 1444.4(352.4) & 0 &  \\
\hline \multirow[c]{9}{*}{\rotatebox{90}{\parbox{2cm}{\centering \exbw \\ (id,N,c)}}} & (01, 5, 0.1) & \bestPerformerCROC 30 & \bestPerformerCROC 8.6(0.1) & \bestPerformerCROC 30 & \bestPerformerCROC 8.5(0.1) & \bestPerformerCROC 30 & \bestPerformerCROC 8.5(0.1) & 30 & 1.6(0.0) & \bestPerformerLMCut \bestPerformer 30 & \bestPerformerLMCut \bestPerformer 1.2(0.0) & 30 & 97.2(1.0) & 30 & 7.9(0.1) & 30 & 626.6(18.8) \\
 & (02, 5, 0.07) & 30 & 40.4(0.3) & 30 & 36.0(0.2) & \bestPerformerCROC 30 & \bestPerformerCROC 35.3(0.3) & 30 & 18.7(0.1) & \bestPerformerLMCut \bestPerformer 30 & \bestPerformerLMCut \bestPerformer 15.2(0.1) & 0 &  & 30 & 732.9(30.6) & 0 &  \\
 & (03, 6, 0.91) & 30 & 173.5(5.1) & \bestPerformerCROC 30 & \bestPerformerCROC 138.0(1.4) & \bestPerformerCROC 30 & \bestPerformerCROC 135.7(1.2) & 30 & 7.5(0.1) & \bestPerformerLMCut \bestPerformer 30 & \bestPerformerLMCut \bestPerformer 3.3(0.0) & 0 &  & 30 & 10.9(0.2) & 0 &  \\
 & (04, 6, 0.16) & 30 & 161.2(1.4) & \bestPerformerCROC 30 & \bestPerformerCROC 142.9(1.0) & \bestPerformerCROC 30 & \bestPerformerCROC 142.1(1.7) & 30 & 63.7(0.5) & \bestPerformerLMCut \bestPerformer 30 & \bestPerformerLMCut \bestPerformer 45.6(0.3) & 0 &  & 27 & 1462.2(79.6) & 0 &  \\
 & (05, 7, 0.01) & 30 & 30.7(0.2) & 30 & 28.1(0.2) & \bestPerformerCROC 30 & \bestPerformerCROC 27.2(0.3) & 30 & 25.8(0.2) & \bestPerformerLMCut \bestPerformer 30 & \bestPerformerLMCut \bestPerformer 12.8(0.1) & 30 & 82.7(1.5) & 30 & 38.5(0.6) & 30 & 451.0(14.5) \\
 & (06, 8, 0.3) & 30 & 300.4(10.9) & \bestPerformerCROC 30 & \bestPerformerCROC 276.0(3.1) & \bestPerformerCROC 30 & \bestPerformerCROC 276.7(4.1) & 30 & 91.6(0.6) & \bestPerformerLMCut \bestPerformer 30 & \bestPerformerLMCut \bestPerformer 48.2(0.4) & 0 &  & 0 &  & 0 &  \\
 & (07, 8, 0.5) & \bestPerformerCROC 30 & \bestPerformerCROC 0.3(0.0) & 30 & 0.5(0.0) & 30 & 0.5(0.0) & \bestPerformerLMCut \bestPerformer 30 & \bestPerformerLMCut \bestPerformer 0.2(0.0) & \bestPerformerLMCut \bestPerformer 30 & \bestPerformerLMCut \bestPerformer 0.2(0.0) & 30 & 2.2(0.1) & 30 & 0.9(0.1) & 30 & 12.1(0.4) \\
 & (08, 8, 0.63) & \bestPerformerCROC \bestPerformer 30 & \bestPerformerCROC \bestPerformer 35.1(0.5) & 30 & 40.5(0.8) & 30 & 41.1(1.0) & \bestPerformerLMCut \bestPerformer 30 & \bestPerformerLMCut \bestPerformer 34.8(0.4) & \bestPerformerLMCut \bestPerformer 30 & \bestPerformerLMCut \bestPerformer 35.5(0.7) & 6 & 1799.2(0.1) & 13 & 1691.9(45.5) & 0 &  \\
 & (09, 8, 0.4) & \bestPerformerCROC 30 & \bestPerformerCROC 177.9(1.4) & 30 & 374.6(3.3) & 30 & 374.2(5.2) & \bestPerformerLMCut \bestPerformer 30 & \bestPerformerLMCut \bestPerformer 14.9(0.1) & 30 & 24.5(0.1) & 0 &  & 30 & 202.3(7.5) & 0 &  \\
\hline \multirow[c]{12}{*}{\rotatebox{90}{\parbox{2cm}{\centering \parcn \\ (f,u)}}} & (0.0, 1) & 30 & 311.2(13.6) & 30 & 183.7(6.9) & \bestPerformerCROC \bestPerformer 30 & \bestPerformerCROC \bestPerformer 1.5(1.4) & 30 & 270.9(8.3) & 30 & 252.5(8.7) & 30 & 126.2(9.1) & \bestPerformerLMCut 30 & \bestPerformerLMCut 103.1(2.4) & 30 & 42.8(3.1) \\
 & (0.0, \(\infty\)) & 30 & 100.7(0.8) & 30 & 52.1(0.9) & \bestPerformerCROC \bestPerformer 30 & \bestPerformerCROC \bestPerformer 0.1(0.0) & 30 & 141.3(1.0) & 30 & 64.7(0.5) & 30 & 47.1(2.5) & \bestPerformerLMCut 30 & \bestPerformerLMCut 42.1(1.8) & 30 & 29.5(1.8) \\
 & (0.2, 1) & 30 & 600.7(3.6) & \bestPerformerCROC 30 & \bestPerformerCROC 234.8(3.1) & 13 & 184.2(3.5) & 30 & 863.4(9.7) & \bestPerformerLMCut 30 & \bestPerformerLMCut 337.1(3.8) & 0 &  & 0 &  & \bestPerformer 30 & \bestPerformer 65.1(11.1) \\
 & (0.2, \(\infty\)) & 30 & 121.7(1.9) & 30 & 91.0(1.3) & \bestPerformerCROC 30 & \bestPerformerCROC 56.9(0.6) & 30 & 185.7(1.2) & \bestPerformerLMCut 30 & \bestPerformerLMCut 79.2(0.6) & 30 & 1537.7(72.8) & 0 &  & \bestPerformer 30 & \bestPerformer 41.7(4.9) \\
 & (0.4, 1) & 30 & 882.8(31.7) & \bestPerformerCROC \bestPerformer 30 & \bestPerformerCROC \bestPerformer 247.5(3.8) & \bestPerformerCROC \bestPerformer 30 & \bestPerformerCROC \bestPerformer 241.8(3.2) & 30 & 564.1(12.8) & \bestPerformerLMCut 30 & \bestPerformerLMCut 308.2(4.0) & 0 &  & 0 &  & 3 & 801.5(536.2) \\
 & (0.4, \(\infty\)) & 30 & 130.9(2.5) & 30 & 98.0(1.8) & \bestPerformerCROC 30 & \bestPerformerCROC 94.0(1.7) & 30 & 188.1(1.3) & \bestPerformerLMCut \bestPerformer 30 & \bestPerformerLMCut \bestPerformer 82.0(1.2) & 0 &  & 0 &  & 3 & 665.9(308.3) \\
 & (0.6, 1) & 30 & 318.4(10.3) & 30 & 186.0(3.6) & \bestPerformerCROC 30 & \bestPerformerCROC 148.3(2.3) & 30 & 255.4(6.9) & \bestPerformerLMCut \bestPerformer 30 & \bestPerformerLMCut \bestPerformer 141.0(2.1) & 0 &  & 0 &  & 0 &  \\
 & (0.6, \(\infty\)) & \bestPerformerCROC \bestPerformer 30 & \bestPerformerCROC \bestPerformer 25.5(0.5) & 30 & 40.0(0.9) & 30 & 40.4(0.9) & \bestPerformerLMCut 30 & \bestPerformerLMCut 39.6(0.3) & 30 & 40.9(0.3) & 26 & 1508.9(68.9) & 0 &  & 0 &  \\
 & (0.8, 1) & 30 & 345.5(7.4) & 30 & 185.3(2.4) & \bestPerformerCROC \bestPerformer 30 & \bestPerformerCROC \bestPerformer 148.2(2.2) & 30 & 267.1(8.0) & \bestPerformerLMCut \bestPerformer 30 & \bestPerformerLMCut \bestPerformer 148.1(2.2) & 0 &  & 0 &  & 0 &  \\
 & (0.8, \(\infty\)) & \bestPerformerCROC \bestPerformer 30 & \bestPerformerCROC \bestPerformer 25.7(0.6) & 30 & 40.4(1.1) & 30 & 40.8(1.0) & \bestPerformerLMCut 30 & \bestPerformerLMCut 39.8(0.3) & 30 & 41.3(0.4) & 29 & 1375.1(89.9) & 0 &  & 0 &  \\
 & (1.0, 1) & 30 & 208.1(6.8) & 30 & 184.6(2.2) & \bestPerformerCROC \bestPerformer 30 & \bestPerformerCROC \bestPerformer 150.3(3.1) & 30 & 226.0(5.4) & \bestPerformerLMCut \bestPerformer 30 & \bestPerformerLMCut \bestPerformer 148.6(1.6) & 0 &  & 0 &  & 0 &  \\
 & (1.0, \(\infty\)) & \bestPerformerCROC \bestPerformer 30 & \bestPerformerCROC \bestPerformer 25.6(0.6) & 30 & 40.5(1.1) & 30 & 41.5(1.1) & \bestPerformerLMCut 30 & \bestPerformerLMCut 39.6(0.3) & 30 & 41.1(0.4) & 30 & 1298.1(55.3) & 0 &  & 0 &  \\
\hline \multirow[c]{9}{*}{\rotatebox{90}{\parbox{2cm}{\centering \tireworld \\ (n,d,c)}}} & (4, 4, 2) & \bestPerformerCROC \bestPerformer 30 & \bestPerformerCROC \bestPerformer 1.6(0.1) & 30 & 2.7(0.1) & 30 & 3.2(0.1) & \bestPerformerLMCut \bestPerformer 30 & \bestPerformerLMCut \bestPerformer 1.8(0.1) & \bestPerformerLMCut \bestPerformer 30 & \bestPerformerLMCut \bestPerformer 1.8(0.1) & 30 & 4.5(0.1) & 30 & 24.2(0.7) & 30 & 24.3(0.8) \\
 & (4, 4, 4) & \bestPerformerCROC \bestPerformer 30 & \bestPerformerCROC \bestPerformer 2.4(0.0) & 30 & 4.6(0.1) & 30 & 5.4(0.1) & \bestPerformerLMCut 30 & \bestPerformerLMCut 3.2(0.0) & \bestPerformerLMCut 30 & \bestPerformerLMCut 3.2(0.0) & 30 & 8.2(0.5) & 30 & 43.0(1.0) & 30 & 36.1(2.0) \\
 & (4, 4, 6) & \bestPerformerCROC \bestPerformer 30 & \bestPerformerCROC \bestPerformer 3.1(0.0) & 30 & 6.6(0.1) & 30 & 7.4(0.1) & \bestPerformerLMCut 30 & \bestPerformerLMCut 4.6(0.1) & \bestPerformerLMCut 30 & \bestPerformerLMCut 4.7(0.1) & 30 & 10.5(0.4) & 30 & 56.0(2.4) & 30 & 42.5(1.5) \\
 & (4, 5, 2) & \bestPerformerCROC \bestPerformer 30 & \bestPerformerCROC \bestPerformer 3.9(0.2) & 30 & 6.3(0.3) & 30 & 7.4(0.3) & \bestPerformerLMCut \bestPerformer 30 & \bestPerformerLMCut \bestPerformer 4.2(0.3) & \bestPerformerLMCut \bestPerformer 30 & \bestPerformerLMCut \bestPerformer 4.2(0.2) & 30 & 18.7(0.8) & 30 & 119.1(5.5) & 30 & 115.3(6.3) \\
 & (4, 5, 4) & \bestPerformerCROC \bestPerformer 30 & \bestPerformerCROC \bestPerformer 5.9(0.1) & 30 & 11.3(0.3) & 30 & 13.4(0.6) & \bestPerformerLMCut 30 & \bestPerformerLMCut 7.5(0.1) & \bestPerformerLMCut 30 & \bestPerformerLMCut 7.6(0.2) & 30 & 32.4(1.0) & 30 & 239.5(9.0) & 30 & 157.0(5.3) \\
 & (4, 5, 6) & \bestPerformerCROC \bestPerformer 30 & \bestPerformerCROC \bestPerformer 7.3(0.3) & 30 & 15.9(0.4) & 30 & 18.6(0.5) & \bestPerformerLMCut 30 & \bestPerformerLMCut 10.3(0.4) & \bestPerformerLMCut 30 & \bestPerformerLMCut 11.1(0.5) & 30 & 41.7(1.0) & 30 & 310.6(14.7) & 30 & 194.5(7.1) \\
 & (4, 6, 2) & \bestPerformerCROC \bestPerformer 30 & \bestPerformerCROC \bestPerformer 9.3(0.5) & 30 & 14.9(0.6) & 30 & 17.8(0.8) & \bestPerformerLMCut \bestPerformer 30 & \bestPerformerLMCut \bestPerformer 10.1(0.6) & \bestPerformerLMCut \bestPerformer 30 & \bestPerformerLMCut \bestPerformer 9.8(0.6) & 30 & 90.9(4.7) & 30 & 840.2(47.1) & 30 & 937.7(45.7) \\
 & (4, 6, 4) & \bestPerformerCROC \bestPerformer 30 & \bestPerformerCROC \bestPerformer 13.9(0.3) & 30 & 27.1(0.8) & 30 & 31.3(0.9) & \bestPerformerLMCut 30 & \bestPerformerLMCut 17.8(0.3) & 30 & 18.5(0.3) & 30 & 187.0(5.9) & 13 & 1656.6(76.8) & 30 & 1281.6(64.0) \\
 & (4, 6, 6) & \bestPerformerCROC \bestPerformer 30 & \bestPerformerCROC \bestPerformer 18.2(0.4) & 30 & 37.7(0.9) & 30 & 44.7(0.9) & \bestPerformerLMCut 30 & \bestPerformerLMCut 24.9(0.7) & 30 & 26.5(0.6) & 30 & 225.1(10.0) & 4 & 1424.3(88.9) & 30 & 1238.9(81.6) \\
\hline \end{tabular}

}
\caption{For the benchmark problems, we show each planner and heuristic's coverage (out of 30) and over the converged runs the mean runtime (secs) with 95\% C.I.
The fastest planner and heuristic per problem are in bold.%
}
\label{tab:extended-results}
\end{table*}

\bibliography{references}

\end{document}